\documentclass{zlab}

\usepackage[T1]{fontenc}
\usepackage{lmodern}

\usepackage{xspace}
\usepackage{graphicx,amsmath,amssymb,hyperref}
\usepackage[authoryear,round]{natbib}
\usepackage{multirow}
\usepackage{xcolor}

\usepackage{url}
\usepackage{nicefrac}
\usepackage{wrapfig}
\usepackage{rotating}
\usepackage{subcaption}
\usepackage{algorithm}
\usepackage{algorithmic}
\usepackage{listings}
\definecolor{lstbg}{HTML}{F7F8FA}
\definecolor{lstframe}{HTML}{C8D0DA}
\definecolor{lstcomment}{HTML}{6B7280}
\definecolor{lstkeyword}{HTML}{1F6FEB}
\definecolor{lststring}{HTML}{116329}
\lstdefinestyle{codestyle}{
    backgroundcolor=\color{lstbg},
    commentstyle=\color{lstcomment}\itshape,
    keywordstyle=\color{lstkeyword}\bfseries,
    stringstyle=\color{lststring},
    basicstyle=\ttfamily\footnotesize,
    breakatwhitespace=false,
    breaklines=true,
    keepspaces=true,
    showspaces=false,
    showstringspaces=false,
    showtabs=false,
    tabsize=2,
    frame=single,
    framesep=6pt,
    framerule=0.5pt,
    rulecolor=\color{lstframe},
    xleftmargin=10pt,
    xrightmargin=4pt,
    aboveskip=8pt,
    belowskip=8pt,
    escapeinside={(*@}{@*)},
    captionpos=b,
}
\definecolor{promptbg}{HTML}{F4F1FA}
\definecolor{promptframe}{HTML}{B7A6D9}
\lstdefinestyle{promptstyle}{
    style=codestyle,
    backgroundcolor=\color{promptbg},
    rulecolor=\color{promptframe},
    basicstyle=\sffamily\footnotesize,
    keywordstyle=\bfseries,
    captionpos=b,
}
\definecolor{cardbg}{HTML}{EDF7EE}
\definecolor{cardframe}{HTML}{8FBF9E}
\lstdefinestyle{cardstyle}{
    style=codestyle,
    backgroundcolor=\color{cardbg},
    rulecolor=\color{cardframe},
    keywordstyle=\bfseries,
    captionpos=b,
}

\definecolor{algkw}{RGB}{150, 150, 150}
\definecolor{algcmd}{RGB}{0, 90, 160}
\definecolor{alglib}{RGB}{140, 50, 140}
\definecolor{algeval}{RGB}{190, 100, 20}

\newcommand{\method}{\textsc{Strategist}}

\hypersetup{
  colorlinks=true,
  urlcolor=magenta,
  linkcolor=link,
  citecolor=link,
  filecolor=black,
  pdfborder={0 0 0}
}

\definecolor{link}{HTML}{307EF1}
\makeatletter
\let\zlab@footnotemark\@footnotemark
\def\@footnotemark{\begingroup\hypersetup{linkcolor=black}\zlab@footnotemark\endgroup}
\makeatother
\providecommand{\equationname}{Equation}
\providecommand{\sectionname}{Section}

\newcommand{\figref}[1]{%
  \figurename~\hyperref[#1]{\textcolor{link}{\ref*{#1}}}%
}
\newcommand{\tabref}[1]{%
  \tablename~\hyperref[#1]{\textcolor{link}{\ref*{#1}}}%
}
\newcommand{\eqrefc}[1]{%
  \equationname~\hyperref[#1]{\textcolor{link}{\ref*{#1}}}%
}
\newcommand{\secrefc}[1]{%
  \sectionname~\hyperref[#1]{\textcolor{link}{\ref*{#1}}}%
}

\newcommand{\github}{\raisebox{-1.3pt}{\includegraphics[height=1.05em]{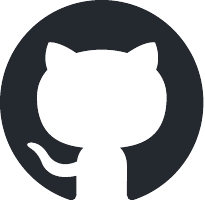}}\xspace}

\makeatletter
\fancypagestyle{firststyle}{
  \fancyhead[L]{}
  \fancyhead[C]{}
  \fancyhead[R]{}
  \fancyfoot[L]{\footerfont \the\correspondingauthor}
  \fancyfoot[R]{}
}
\makeatother

\makeatletter
\DeclareRobustCommand\bfseries{%
  \not@math@alphabet\bfseries\mathbf
  \fontseries\bfdefault\selectfont
  \sffamily
}
\makeatother
\DeclareTextFontCommand{\textbf}{\bfseries\sffamily}
\title{\centering {\fontsize{18}{24}\selectfont \method{}}: Evolving Reasoning with Agents}

\author{
    \vspace{.2cm}
    \parbox{\textwidth}{\centering
        Taiming Lu\textsuperscript{1} \quad
        Qi Xu\textsuperscript{2} \quad
        Zhigang Hua\textsuperscript{2} \quad
        Zhuang Liu\textsuperscript{1}
    }
    \\
    \vspace{-.05cm}
    {\normalfont\fontsize{11}{15}\selectfont {\textsuperscript{1}Princeton University}\hspace{.1cm}}
    {\normalfont\fontsize{11}{15}\selectfont {\textsuperscript{2}Meta}}
}

\makeatletter
\renewcommand{\abscontent}{}%
\makeatother

\newenvironment{abstractblock}{%
  {\centering\large\bfseries Abstract\par}
  \vspace{0em}
  \begin{list}{}{%
      \setlength{\leftmargin}{3.0em}
      \setlength{\rightmargin}{3.0em}
      \setlength{\topsep}{0pt}
      \setlength{\parsep}{0pt}
  }
  \item[]
}{%
  \end{list}
  \par\normalfont\vspace{1em}
}

\begin{document}

\begingroup
\makeatletter
\let\raggedright\centering
\makeatother

\maketitle
\endgroup

\newcommand{\abstractcontent}{%
A model is only half of an AI agent system. The other half is the inference strategy: the reasoning procedure that turns model calls into a response. Today that second half is written by hand, and the experience of building one strategy never carries to the next. We introduce \method{}, a meta-agent that evolves inference strategies across tasks. Each strategy is written as a blueprint, executable code that other strategies can reuse, and together the blueprints form a library. Given a task and a small development set, \mbox{\method{}} retrieves related blueprints, writes candidate strategies, tests them on real samples, and revises them from the results. Every candidate then enters the library, so later tasks build on earlier discoveries. On 30 benchmarks, \method{} beats the strongest baseline on 24 and matches its average accuracy ($74.7\%$ vs.\ $74.0\%$) at 39\% lower inference cost. Its strategies also transfer to unseen tasks and other models without further search. The part of an AI system that is not improved on its own now does.{\parfillskip=0pt\par}
}

\vspace{-0.3cm}

\begin{abstractblock}
\abstractcontent
\end{abstractblock}

\vspace{0.6em}
{\centering \github \href{https://github.com/zlab-princeton/Strategist}{\textcolor{black}{Code}}\par}
\vspace{-0.15em}

\noindent\begin{minipage}{\linewidth}
\centering
\includegraphics[width=\linewidth]{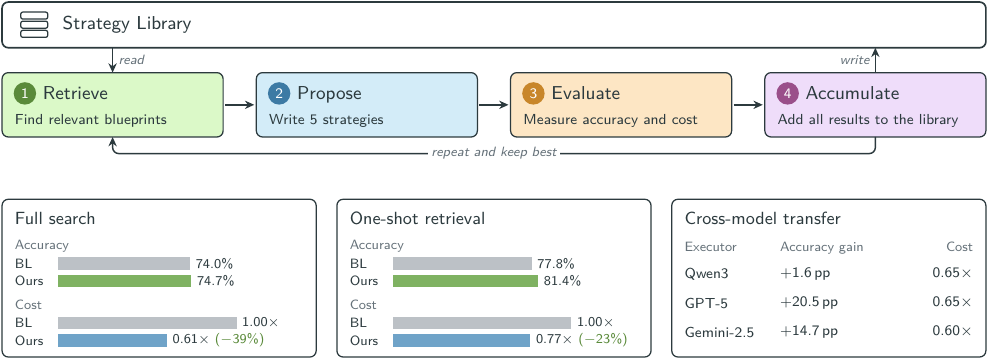}
\captionof{figure}{\textbf{\method{} turns strategy design into a search that accumulates knowledge.} \textit{Top:} in each round, an LLM writes new inference strategies as code, runs them, and stores every one, with its measured accuracy and cost, in a library shared across tasks, so later tasks build on what earlier ones found. The models stay frozen; only the strategies change. \textit{Bottom:} the search matches the strongest baseline (BL) at lower cost, and its strategies work on new tasks and new models without further search. Cost ratios are per-benchmark averages; transfer costs are totals in USD.{\parfillskip=0pt\endgraf}}
\label{fig:teaser}
\end{minipage}

\newpage

\section{Introduction}
\label{sec:intro}

\begin{figure}[b]
    \centering
    \vspace{-6pt}
    \includegraphics[width=\linewidth]{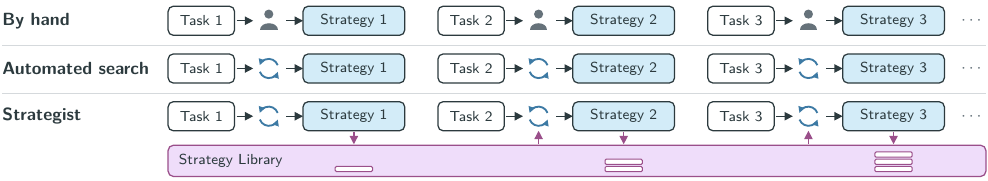}
    \caption{\textbf{Strategy design is per task; \method{} makes it cumulative.} By hand and in prior automated search, nothing carries between tasks. \method{} keeps every strategy in a library that the next task starts from.}
    \label{fig:cumulative}
\end{figure}

A model is only half of an AI system. The other half is how that model is put to work, and it matters just as much. Everything from the prompt to the reasoning procedure decides how it behaves. Today that second half is designed by people, fixed, one system at a time. Someone writes the prompt, tries it, adjusts it, and once the system works, the design is left as it is. It is the part of the system that never improves on its own.

There is no reason it has to stay that way. The prompts and reasoning procedures that put a model to work are text and code, and agents themselves have become good at writing both. What humans do to design (write, try, adjust) such \textit{inference strategies} is the natural agentic loop. Such loops have already been shown to evolve programs far beyond their starting point~\citep{novikov2025alphaevolve}, and AI scientists now run the same loop end to end~\citep{lu2024aiscientist, gottweis2025coscientist,ai-scientistv2}. Designing an inference strategy is the same work with a different object. So why not evolve agentic reasoning itself the same way? Run on far more cases than a person could try, 
the loop could outdo what people build by hand.

Recent work has begun to automate this loop: searching over strategy designs, as raw code or as assemblies of fixed components~\citep{hu2024adas, zhang2024aflow, shang2024agentsquare, zhang2025evoflow}; training controllers that route each input to an existing strategy~\citep{arc2025, yue2025masrouter, bingreeky2025maas, daao2025}; or optimizing one strategy per task by training, evolution, or iterative refinement~\citep{gao2023strategyllm, autoflow2025, zhang2026flowsteer, flowreasoner2025, sew2025}. All of them beat hand-designed strategies on individual benchmarks, and all of them stop there. Searching or routing over a fixed set of strategies can never discover a new one. Trained optimizers need to start over for every task. These methods make the design automatic. We want to make it cumulative (\figref{fig:cumulative}).

We start from the representation. Evolving strategies needs a form open enough to express any reasoning procedure and structured enough for later searches to reuse and combine. We introduce the \emph{Strategy Blueprint}: an executable program with one shared code interface, so any blueprint can serve as a component inside another. New designs are built from old ones, and the library grows as it is used.

On this representation we build \method{}, a meta-agent that does not answer questions but designs the strategies that do. It runs the write, try, adjust loop at scale, with frozen models and no training. Given a task and a small development set, it writes candidate strategies as blueprints, runs them on real samples, reads the traces, and revises them against measured accuracy and cost. Before each round it queries a shared library for relevant blueprints and for diverse ones as inspiration. 
After each round every candidate
goes back into the library with its
measurements and experience notes.

On 30 benchmarks spanning math, code, commonsense, multi-hop QA, knowledge, logic, and recommendation, we compare \method{} against 18 baselines.
\method{} wins on 24 of 30, matching the strongest baseline's average accuracy ($74.0\%\!\to\!74.7\%$) at 39\% lower cost. The library pays off on its own: on 10 held-out benchmarks, retrieval with no search stays within 2 points of the strongest baseline on all 10 and beats it on 8 ($77.8\%\to81.4\%$). The same strategies, moved to GPT-5 and Gemini-2.5 without changing a line, still beat the baselines there. Each task solved leaves the next one a better place to start. Until now, only the model in an AI system could learn. Now the rest of it can too.

\begin{figure}[t]
    \centering
    \includegraphics[width=0.9\linewidth]{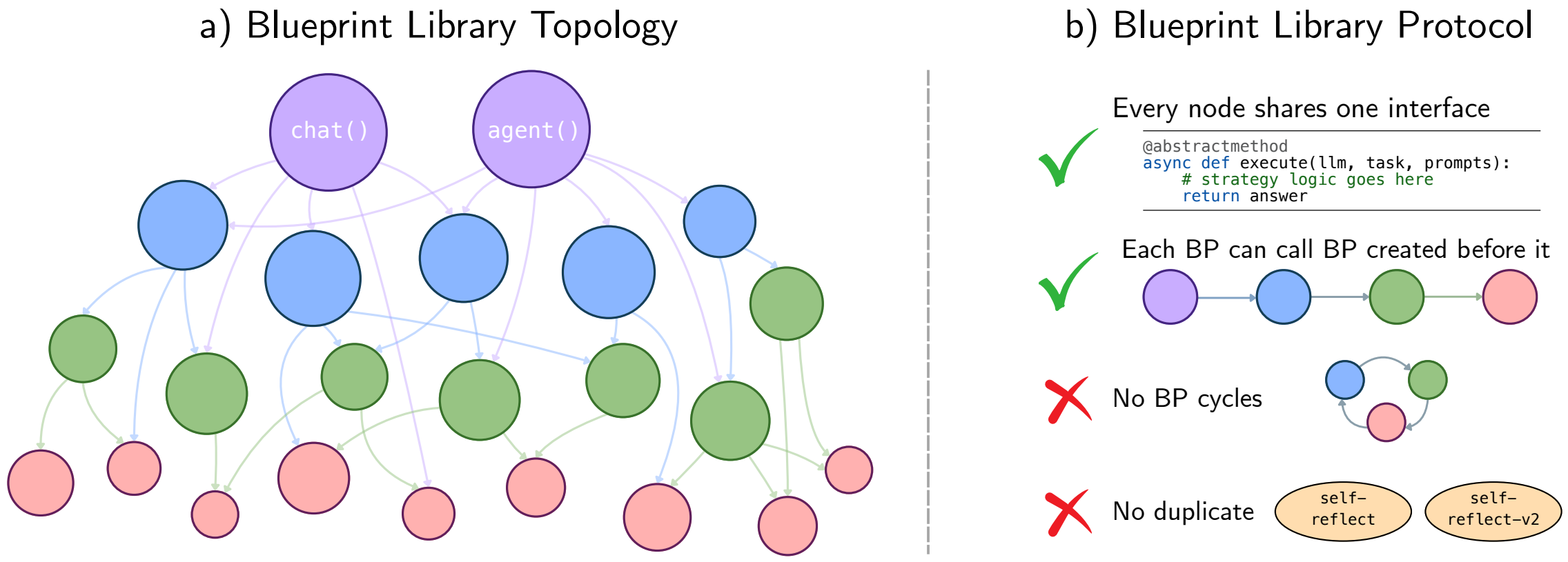}
    \caption{\textbf{The Strategy Blueprint is a freeform graph that grows through the library.} (a)~Two foundations (\texttt{chat} and \texttt{agent}) sit at the top. Strategies compose from them and from each other, forming a directed graph whose levels emerge naturally from composition depth. (b)~Four protocols govern the graph: every node shares one abstract interface, each node can only call nodes created before it, no cycles are allowed, and no duplicates should exist.}
    \label{fig:blueprint}
    \vspace{-.3cm}
\end{figure}

\newpage

\section{The Strategy Blueprint}
\label{sec:blueprint}

Every inference strategy, ultimately, is a program. How it is represented determines what an automated designer can build, test, and reuse. Existing work represents strategies as computation graphs~\citep{wu2024stateflow, zhuge2024gptswarm} and declarative pipelines~\citep{khattab2024dspy, survey2025acg}, which are composable but can only express pre-defined patterns, or as monolithic code~\citep{hu2024adas}, which can express anything but produces isolated strategies. We introduce the \emph{Strategy Blueprint}: a reasoning flow that specifies how to solve a task (which models to call, in what order, at which tier, and how) using the open interface of code. Every blueprint exposes the same interface, so any can call any other, and a library can grow by composition.

\begin{wrapfigure}[19]{r}{0.4\linewidth}
    \vspace{-1.6em}
    \centering
    \includegraphics[width=\linewidth]{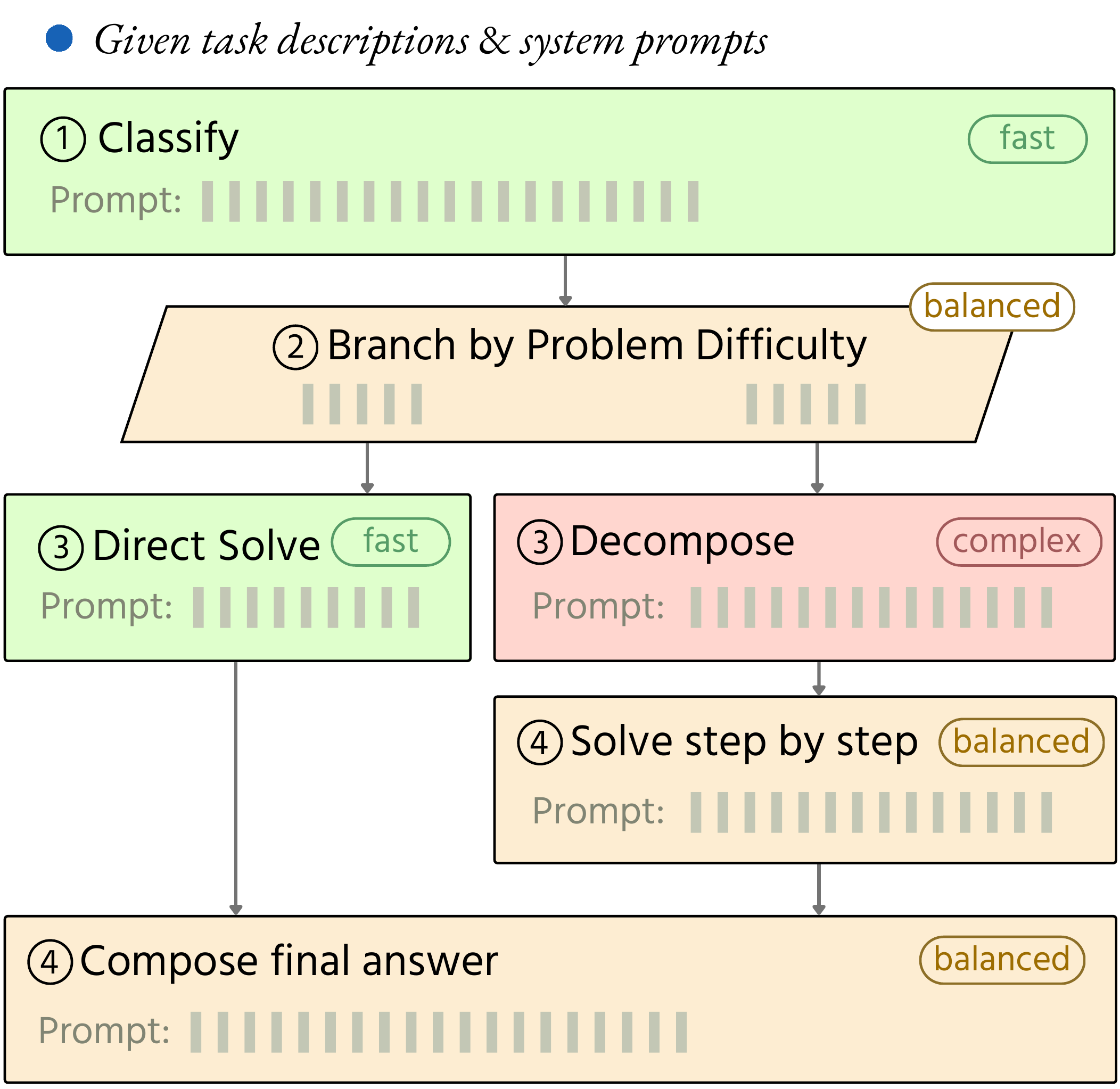}
    \vspace{-1.3em}
    \caption{\textbf{Strategy blueprint captures reasoning structure.} This simplified blueprint example defines the steps, control flow, and the model tier for each step. Prompts are not specified; filling in produces a runnable strategy.}
    \label{fig:blueprint_example}
    \vspace{-1.5em}
\end{wrapfigure}

\paragraph{Foundations.}
A blueprint is a Python program that captures the \textit{structure} of a reasoning strategy without the task-specific prompts. For example, one blueprint defines a loop: generate three candidate answers, review each, then majority-vote. Filling in prompts produces a runnable \emph{strategy}. It has one interface, \texttt{execute(llm, task) -> str}, and is built from two foundations: \texttt{chat} (single- or multi-turn conversation) and \texttt{agent} (with scaffold and tools). See \figref{fig:blueprint_example} for an example.

\paragraph{Multi-tier model routing.}
We allow strategies to route across three model tiers (\texttt{fast}, \texttt{balanced}, \texttt{complex}), each in two modes (\texttt{nothink} and \texttt{thinking}), mirroring the tiered pricing and effort levels of common model providers. Not every reasoning step needs the same intelligence, so each step can target a different tier. This internal routing is what allows a single strategy to be accurate and cost-efficient at the same time.

\paragraph{The Blueprint Library.}
Blueprints live in a shared library that grows over time. They are added along with their measured performance and experience notes on when it works and when it does not. Because every blueprint exposes the same interface, a new blueprint can use any existing one as a component, so the library compounds: early blueprints build on the two foundations, later ones compose from those, and depth grows naturally (\figurename~\hyperref[fig:blueprint]{\textcolor{link}{\ref*{fig:blueprint}a}}). Four protocols govern the library (\figurename~\hyperref[fig:blueprint]{\textcolor{link}{\ref*{fig:blueprint}b}}): a shared interface, no forward calls, no cycles, and no duplicates.

\begin{figure}[t]
\vspace{-1cm}
\centering
\includegraphics[width=.9\linewidth,trim=0.3cm 0cm 0.3cm 0cm]{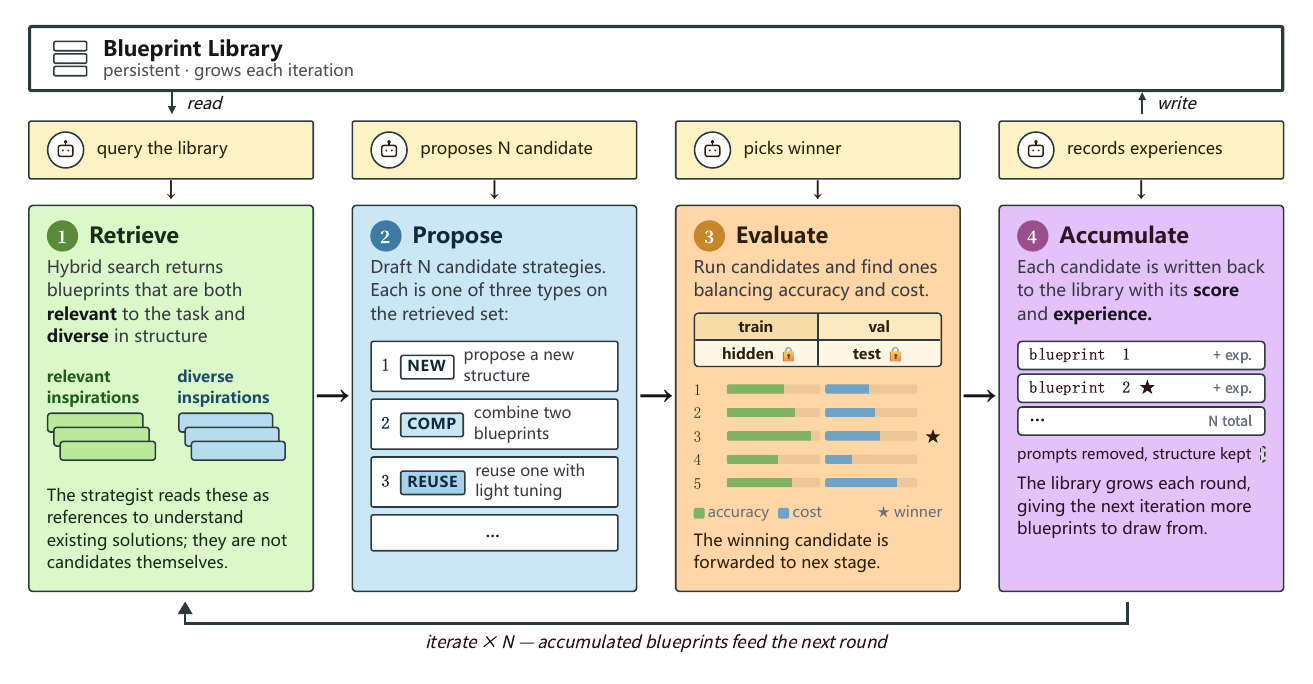}
\vspace{-.4cm}
\caption{\textbf{The \method{} four-stage iterative pipeline.} Each iteration: (1)~\textit{Retrieve} relevant blueprints and diverse inspiration from the Library; (2)~\textit{Propose} $K$ candidates as a mix of \emph{new}, \emph{composed}, and \emph{reuse} blueprints paired with task-specific prompts; (3)~\textit{Evaluate} each candidate, seeing full traces on the training set and aggregated scores on validation; (4)~\textit{Accumulate} the underlying blueprints into the library with measurement cards and experience notes.}
\label{fig:pipeline}
\vspace{-.3cm}
\end{figure}

\section{\method{}}
\label{sec:meta}

Given a task and a small development set, \method{} iteratively retrieves relevant blueprints, proposes $K{=}5$ candidates, evaluates them on real samples, and accumulates the results back into the library (\figref{fig:pipeline}). After $N$ iterations, the candidate that beats BL on both the validation and hidden splits by the largest hidden-set margin is returned, or BL itself if none does (Algorithm~\ref{alg:loop}).

\paragraph{Step 1: Retrieve.}
The strategist queries the Blueprint Library with a short text prompt. The search is hybrid (TF--IDF plus embedding similarity). Each blueprint carries traits, a measurement card, and experience notes written by previous strategists; the search indexes over all of these, so the strategist finds blueprints by measured behavior and prior analyses, not just structural similarity. Beyond relevant results, the library returns \emph{adjacent inspiration} (blueprints from other domains or reasoning patterns) to inspire new designs.

\paragraph{Step 2: Propose.}
Using the retrieved blueprints and inspiration, the strategist designs new blueprints or composes from existing ones, each paired with task-specific prompts to form candidate strategies. Each blueprint must be one of three kinds: \emph{new} (a blueprint with no library backbone), \emph{composed} (a blueprint that wraps or extends an existing one), or \emph{reuse} (directly applies an existing library blueprint with new prompts). Every iteration includes at least one candidate of each kind, 
balancing exploration and exploitation.

\paragraph{Step 3: Evaluate.}
The system operates at two levels. At the \emph{agent level} (inner loop), each candidate strategy processes inputs directly, and the strategist observes the full reasoning trace, execution status, and scores. At the \emph{meta level} (outer loop), the strategist uses these observations to improve its proposals. The data is split into four parts. The \emph{training set}: strategies run on these samples and the strategist sees both the scores and the full agent traces. The \emph{validation set}: strategies also run on these samples, but the strategist sees only the aggregated scores; this mirrors standard train/validation practice and prevents overfitting to the training signal. The \emph{hidden set}: used to select the winning strategy across all iterations, preventing overfitting to any single iteration. The \emph{test set}: held out entirely from the search loop and used only for final confirmation.

\paragraph{Step 4: Accumulate.}
The library stores blueprints, not strategies: it retains the reusable structure while discarding task-specific prompts. This is what enables cross-task transfer. After each iteration, the strategist writes an analysis of each candidate's structural properties and adds the underlying blueprints to the library with their measurement cards and experience notes, with deduplication.

\section{Experiments}
\label{sec:experiments}

\paragraph{Benchmarks.}
We evaluate on 40 benchmarks spanning seven categories (\tabref{tab:benchmarks}). From these, 30 randomly selected benchmarks form the main results and 10 are held out for one-shot retrieval. Each benchmark is split into training, validation, and hidden sets of up to 100 samples each, plus a fixed 200-sample test set.{\parfillskip=0pt\par}

\begin{table}[!b]
\vspace{-8pt}
\centering
\footnotesize
\setlength{\tabcolsep}{3.5pt}
\renewcommand{\arraystretch}{1.15}
\begin{tabular}{@{}p{64.4pt}p{116.8pt}p{145.8pt}p{\dimexpr\linewidth-327pt-21pt\relax}@{}}
\toprule
\multirow{3}{*}{Math} & AQuA-RAT~\citep{ling2017aqua} & Gaokao-Math~\citep{zhong2023agieval} & GSM8K~\citep{cobbe2021gsm8k} \\
 & AIME 2024~(MAA,~\citeyear{aime2024}) & MATH500~\citep{hendrycks2021math} & ASDiv~\citep{miao2020asdiv} \\
 & SVAMP~\citep{patel2021svamp} & MinervaM~\citep{lewkowycz2022minerva} &  \\
\midrule
\multirow{4}{*}{Commonsense} & StrategyQA~\citep{geva2021strategyqa} & CommonsenseQA~\citep{talmor2019csqa} & PIQA~\citep{bisk2020piqa} \\
 & HellaSwag~\citep{zellers2019hellaswag} & OpenBookQA~\citep{mihaylov2018openbookqa} & SIQA~\citep{sap2019siqa} \\
 & COPA~\citep{roemmele2011copa} & WinoGrande~\citep{sakaguchi2020winogrande} & SciQ~\citep{welbl2017sciq} \\
 & BoolQ~\citep{clark2019boolq} &  &  \\
\midrule
\multirow{2}{*}{Knowledge} & TruthfulQA~\citep{lin2022truthfulqa} & SQuAD v2~\citep{rajpurkar2018squad2} & MMLU-STEM-Hard \\
 & TriviaQA~\citep{joshi2017triviaqa} & NQ-Open~\citep{kwiatkowski2019nq} & MMLU~\citep{hendrycks2021mmlu} \\
\midrule
\multirow{2}{*}{Multi-hop} & QUALITY~\citep{pang2022quality} & 2WikiMQA~\citep{ho2020twowiki} & DROP~\citep{dua2019drop} \\
 & MuSiQue~\citep{trivedi2022musique} & HotpotQA~\citep{yang2018hotpotqa} &  \\
\midrule
\multirow{3}{*}{Logic} & LSAT-RC~\citep{zhong2023agieval} & SAT-Math~\citep{zhong2023agieval} & MedQA~\citep{jin2021medqa} \\
 & ANLI-R3~\citep{nie2020anli} & LSAT-AR~\citep{zhong2023agieval} & ReCLor~\citep{yu2020reclor} \\
 & SAT-EN~\citep{zhong2023agieval} & LSAT-LR~\citep{zhong2023agieval} & BBH~\citep{suzgun2023bbh} \\
\midrule
Code & MBPP+~\citep{liu2023evalplus} &  &  \\
\midrule
Recommendation & Amazon-Rec~\citep{liu2025onerecthink} &  &  \\
\bottomrule
\end{tabular}
\caption{\textbf{40 benchmarks across seven task categories.} We randomly select 30 for the main iterative search and hold out the remaining 10 for one-shot retrieval. Each benchmark is split into training, validation, hidden, and test sets.\parfillskip=0pt}
\label{tab:benchmarks}

\vspace{14pt}
\footnotesize
\setlength{\tabcolsep}{3.5pt}
\begin{tabular}{@{}p{64.4pt}p{116.8pt}p{145.8pt}p{\dimexpr\linewidth-327pt-21pt\relax}@{}}
\toprule
\multirow{4}{*}{Prompting} & Monte-Carlo vote & Chain-of-thought~\citep{wei2022cot} & Self-discover~\citep{zhou2024selfdiscover} \\
 & Direct thinking & Tree of thoughts~\citep{yao2023tot} & Self-refine~\citep{madaan2023selfrefine} \\
 & Shortlist vote & Self-consistency~\citep{wang2023selfconsistency} & Reflexion~\citep{shinn2023reflexion} \\
 & Direct & Least-to-most~\citep{zhou2023leasttomost} & CoVe~\citep{dhuliawala2023cove} \\
\midrule
\multirow{2}{*}{Agentic} & Agent with code & Multi-agent debate~\citep{du2024debate} & MetaGPT~\citep{hong2023metagpt} \\
 & Vanilla agent & Chain-of-agents~\citep{zhang2024chainofagents} & AutoGen~\citep{wu2023autogen} \\
\bottomrule
\end{tabular}
\caption{\textbf{Baselines.} Prompting methods call the model directly; agentic methods add tool use or multiple interacting agents. Each baseline runs at all six tier/mode configurations, and we compare against the strongest training split cell.}
\label{tab:baselines}
\vspace{20pt}
\end{table}

\paragraph{Models.}
The task models consist of three tiers: Qwen3-0.6B (fast), Qwen3-1.7B (balanced), and Qwen3-4B (complex), each run with thinking off (nothink) or on (thinking), giving six model configs. The strategist is GPT-6, a separate frozen LLM accessed through the Codex CLI; it writes strategies but never answers questions.{\parfillskip=0pt\par}

\paragraph{Baselines.}
We compare against 18 baselines (\tabref{tab:baselines}), all using the same task models without internet access. They span from direct prompting to multi-agent workflows, and each runs at all six model configurations, giving $18\times6=108$ cells per benchmark. Our baseline (BL) is the cell most accurate on a benchmark's training split, scored on the test set. This mirrors deployment: a method is chosen on the small sample at hand, then applied to a larger unseen sample from the same distribution. Our strategy is likewise chosen without seeing the test set.{\parfillskip=0pt\par}

\paragraph{Metrics.}
We report accuracy and weighted cost. Cost is token usage (input and output) weighted by tier (fast $= 1\times$, balanced $= 2.83\times$, complex $= 6.67\times$, matching the 0.6B/1.7B/4B parameter ratio), since per-token floating-point operations (FLOPs) scale linearly with the parameter count~\citep{kaplan2020scaling}.{\parfillskip=0pt\par}

\begin{figure}[!t]
\centering
\includegraphics[width=.9\linewidth, trim=0cm 1.5cm 0cm 0cm, clip]{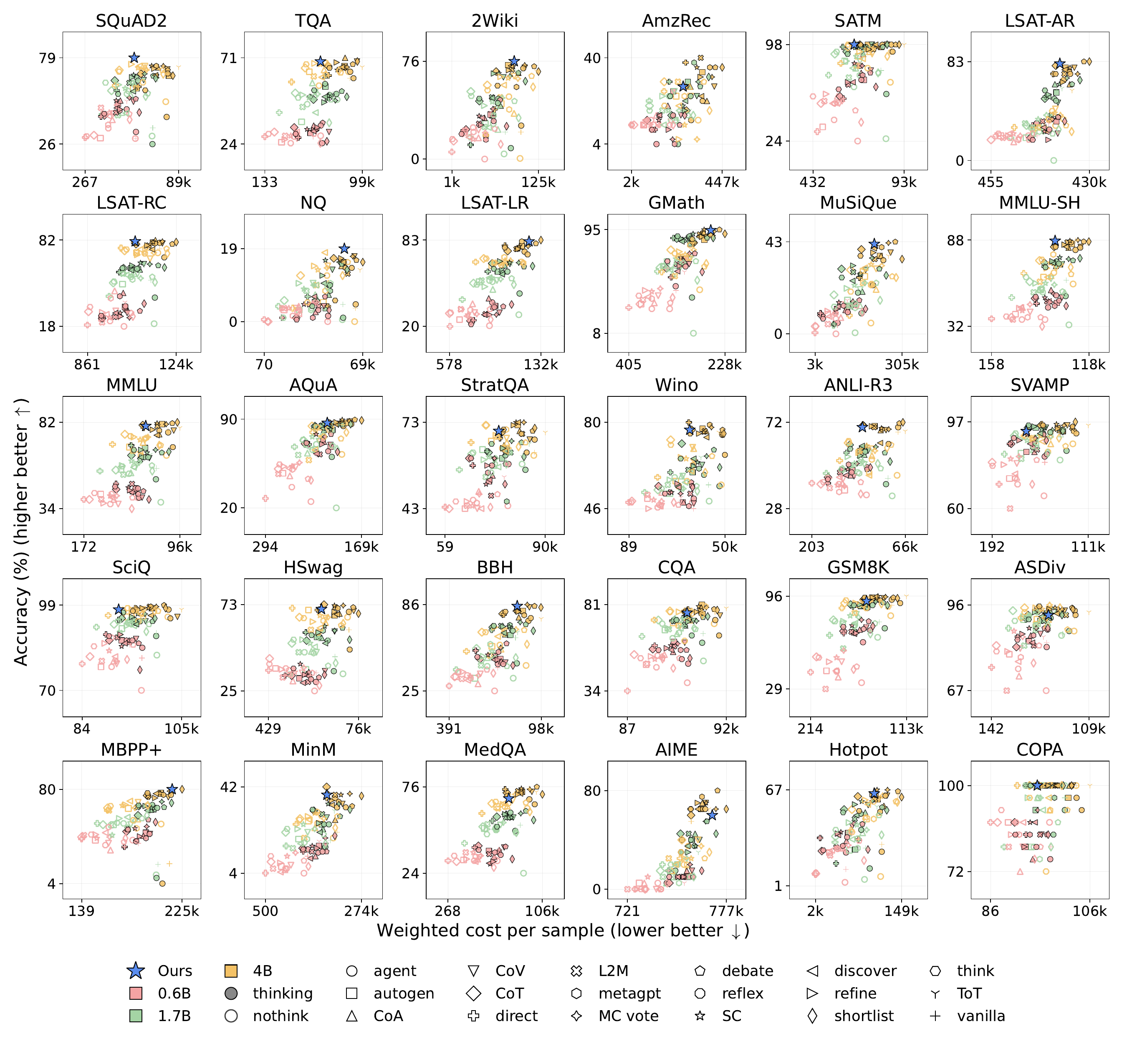}
\caption{\textbf{Ours among all 108 baseline cells (accuracy vs weighted cost per sample, test set).} Each small marker is one baseline cell (a baseline strategy at one model tier and mode); the star is \method{}'s strategy. BL, the comparison point, is the cell chosen on the training split (\figref{fig:avg_bars}). The highest marker in a panel is the best of 108 cells \emph{chosen by looking at the test results}, which no method can do in advance; it is shown for context, not as the comparison point. Appendix~\figref{fig:cost_acc_train} shows the same view on the training split, where the bias runs the other way. Against BL, Ours is more than 2\% cheaper on 26 of 30 benchmarks and within 2 accuracy points or better on 28.}
\label{fig:cost_acc_main}
\end{figure}

\begin{figure}[!t]
\centering
\includegraphics[width=\linewidth, trim=0cm 1.2cm 0cm 0cm]{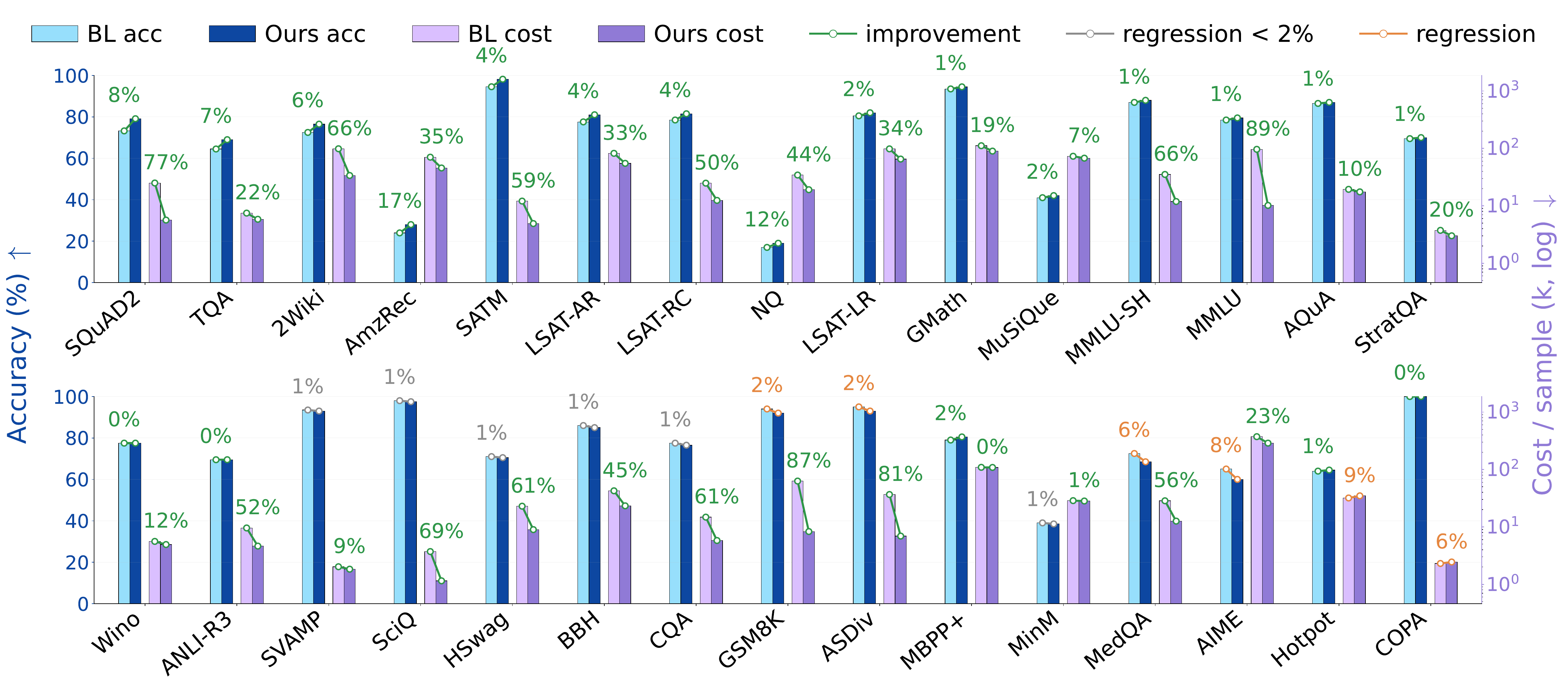}
\caption{\textbf{Per-benchmark head-to-head: \method{} vs the baseline (BL) on the 30 main benchmarks.} For each benchmark, test accuracy (\%, left axis) and weighted cost per sample (log scale, right axis) of Ours and of BL. BL is the baseline cell (one of the $18\times6$ baseline--configuration pairs) that is most accurate on the benchmark's \emph{training} split, scored on the test set; \method{}'s strategy is also chosen without seeing the test set, so this is the like-for-like comparison behind the win counts. Numbers above each pair are relative changes (green: improvement; grey: regression under 2\%; orange: larger regression).}
\label{fig:avg_bars}
\end{figure}


\newpage
\section{Results}
\label{sec:results}

\definecolor{wincol}{RGB}{0, 120, 60}
\definecolor{tiecol}{RGB}{100, 100, 100}
\definecolor{losecol}{RGB}{220, 140, 100}

We run the full \method{} loop on each of the 30 main benchmarks. BL is the cell of the $18 \times 6$ baseline matrix that is most accurate on the benchmark's training split; all methods share the same executors, extractor, and 200-sample test split. A strategy wins if it is better than BL on accuracy (by more than 2 points) or on cost (by more than 2\%) and worse on neither. \method{} wins on 24 of 30 benchmarks, ties on 2 (MBPP+, MinervaM), trades accuracy for cost on 2 (MedQA, AIME) and loses on 2 (HotpotQA and COPA, at equal accuracy but 9\% and 6\% higher cost); it cuts weighted cost by more than 2\% on 26 of 30 (\figref{fig:avg_bars}; \figref{fig:cost_acc_main} places Ours among all 108 baseline cells; per-benchmark numbers in Appendix~\ref{app:main_table}). On average, \method{} matches BL's accuracy ($74.0\%\to74.7\%$) while reducing weighted cost by $39\%$ (a mean per-benchmark cost ratio of $0.61\times$ relative to BL).

\begin{table}[!b]
\centering
\begin{minipage}[t]{0.48\linewidth}
\centering
\footnotesize
\setlength{\tabcolsep}{4pt}
\begin{tabular}{lcccc}
\toprule
& \multicolumn{2}{c}{Test acc (\%)} & \multicolumn{2}{c}{$\Delta$ vs baseline} \\
\cmidrule(lr){2-3} \cmidrule(lr){4-5}
Benchmark & Ours & BL & $\Delta$Acc & $\Delta$Cost \\
\midrule
ReCLor      & 83.5 & 85.0 & \textcolor{tiecol}{$\downarrow$1.8\%} & \textcolor{wincol}{$\downarrow$34.2\%} \\
SAT-EN      & 86.5 & 86.5 & \textcolor{tiecol}{0.0\%} & \textcolor{wincol}{$\downarrow$55.6\%} \\
TriviaQA    & 50.5 & 46.0 & \textcolor{wincol}{$\uparrow$9.8\%} & \textcolor{wincol}{$\downarrow$60.0\%} \\
MATH500     & 94.5 & 92.5 & \textcolor{wincol}{$\uparrow$2.2\%} & \textcolor{wincol}{$\downarrow$22.6\%} \\
BoolQ       & 85.5 & 73.5 & \textcolor{wincol}{$\uparrow$16.3\%} & \textcolor{losecol}{$\uparrow$3103\%} \\
SIQA        & 77.5 & 73.5 & \textcolor{wincol}{$\uparrow$5.4\%} & \textcolor{losecol}{$\uparrow$337\%} \\
OpenBookQA  & 94.5 & 93.0 & \textcolor{wincol}{$\uparrow$1.6\%} & \textcolor{wincol}{$\downarrow$49.2\%} \\
PIQA        & 89.0 & 87.5 & \textcolor{wincol}{$\uparrow$1.7\%} & \textcolor{wincol}{$\downarrow$64.5\%} \\
QUALITY     & 79.5 & 78.5 & \textcolor{wincol}{$\uparrow$1.3\%} & \textcolor{losecol}{$\uparrow$7.4\%} \\
DROP        & 72.5 & 61.5 & \textcolor{wincol}{$\uparrow$17.9\%} & \textcolor{losecol}{$\uparrow$92.4\%} \\
\bottomrule
\end{tabular}
\vspace{0.4em}
\caption{\textbf{One-shot library retrieval.} Top-5 blueprints retrieved from the final strategies, one picked on development questions. On 10 held-out benchmarks, $\Delta$Acc and $\Delta$Cost are relative changes vs.\ the strongest baseline ($(\text{Ours}-\text{BL})/\text{BL}$). Full results in Appendix~\ref{app:oneshot_full}.}
\label{tab:oneshot}
\end{minipage}\hfill
\begin{minipage}[t]{0.50\linewidth}
\centering
\footnotesize
\setlength{\tabcolsep}{3pt}
\begin{tabular}{lcc@{\hspace{8pt}}cc}
\toprule
 & \multicolumn{2}{c}{GPT-5} & \multicolumn{2}{c}{Gemini-2.5} \\
\cmidrule(lr){2-3} \cmidrule(lr){4-5}
Benchmark & $\Delta$Acc & $\Delta$Cost & $\Delta$Acc & $\Delta$Cost \\
\midrule
AQuA-RAT & \textcolor{wincol}{$\uparrow$189\%} & \textcolor{wincol}{$\downarrow$86.5\%} & \textcolor{wincol}{$\uparrow$27.5\%} & \textcolor{wincol}{$\downarrow$51.6\%} \\
2WikiMQA & \textcolor{wincol}{$\uparrow$166\%} & \textcolor{wincol}{$\downarrow$50.8\%} & \textcolor{wincol}{$\uparrow$260\%} & \textcolor{wincol}{$\downarrow$72.0\%} \\
HotpotQA & \textcolor{wincol}{$\uparrow$96.2\%} & \textcolor{wincol}{$\downarrow$7.5\%} & \textcolor{wincol}{$\uparrow$132\%} & \textcolor{wincol}{$\downarrow$52.2\%} \\
MBPP+ & \textcolor{wincol}{$\uparrow$46.0\%} & \textcolor{wincol}{$\downarrow$12.3\%} & \textcolor{wincol}{$\uparrow$15.8\%} & \textcolor{wincol}{$\downarrow$23.7\%} \\
SQuAD2 & \textcolor{wincol}{$\uparrow$28.7\%} & \textcolor{wincol}{$\downarrow$61.5\%} & \textcolor{losecol}{$\downarrow$4.7\%} & \textcolor{wincol}{$\downarrow$57.9\%} \\
NQ-Open & \textcolor{wincol}{$\uparrow$35.0\%} & \textcolor{wincol}{$\downarrow$22.8\%} & \textcolor{wincol}{$\uparrow$118\%} & \textcolor{wincol}{$\downarrow$21.8\%} \\
TruthfulQA & \textcolor{wincol}{$\uparrow$13.7\%} & \textcolor{wincol}{$\downarrow$71.9\%} & \textcolor{wincol}{$\uparrow$9.4\%} & \textcolor{wincol}{$\downarrow$61.9\%} \\
ASDiv & \textcolor{tiecol}{$\uparrow$2.2\%} & \textcolor{losecol}{$\uparrow$163\%} & \textcolor{tiecol}{$\uparrow$2.1\%} & \textcolor{wincol}{$\downarrow$33.1\%} \\
HellaSwag & \textcolor{tiecol}{$\uparrow$0.5\%} & \textcolor{losecol}{$\uparrow$16.9\%} & \textcolor{tiecol}{$\uparrow$0.5\%} & \textcolor{wincol}{$\downarrow$12.3\%} \\
MinervaM & \textcolor{tiecol}{$\uparrow$2.1\%} & \textcolor{losecol}{$\uparrow$4.8\%} & \textcolor{losecol}{$\downarrow$9.4\%} & \textcolor{wincol}{$\downarrow$25.9\%} \\
\bottomrule
\end{tabular}
\vspace{0.4em}
\caption{\textbf{Cross-model transfer.} Strategies discovered on Qwen3, executed directly on GPT-5 and Gemini-2.5 without modification. $\Delta$Acc and $\Delta$Cost are relative changes vs.\ the strongest baseline re-executed on the same backend.
Full numbers in Appendix~\ref{app:transfer_full}.}
\label{tab:transfer}
\end{minipage}
\end{table}

The largest accuracy gains come with large cost cuts: SQuAD v2 ($73.2\%\to79.1\%$ at $0.23\times$ cost), TruthfulQA ($+4.5$ points), 2WikiMQA ($+4.0$ at $0.34\times$), Amazon-Rec ($+4.0$), SAT-Math ($+3.6$ at $0.41\times$), LSAT-AR ($+3.4$) and LSAT-RC ($+3.0$ at $0.50\times$). On benchmarks where the models are near their accuracy ceiling (e.g., MMLU, GSM8K, ASDiv, SciQ, MMLU-STEM-Hard), \method{} keeps accuracy within 2 points of BL and cuts cost instead, by 66--89\%. 
Of the 24 wins, most come from strategist-designed strategies and and some are from simpler baseline scripts, found during search, that match BL's accuracy at a lower cost.
\method{} is not Pareto-improving everywhere: on MedQA and AIME~2024 it is cheaper ($0.44\times$, $0.77\times$) but 4--5 points less accurate, and on HotpotQA and COPA it matches accuracy at slightly higher cost.

\newpage
\paragraph{One-shot library retrieval.}
The 30-benchmark main run grew the library to $1{,}865$ blueprints. We test if a future task can skip the full search: on 10 held-out benchmarks, the strategist queries the final strategies of the main searches (not the full library) once, instantiates the top-5 retrieved blueprints for the task, measures each on up to 300 development questions, and runs the chosen one (the most accurate candidate cheaper than BL if within 2 points of the best, otherwise the best) on the test set, with no search. It stays within 2 points of BL's accuracy on all 10 and above it on 8 ($77.8\%\to81.4\%$ on average) and wins on 6 (\tabref{tab:oneshot}); on BoolQ, SIQA, and DROP it is more accurate but costlier than their very cheap BL ($1.0$--$4.9$k per sample), and on QUALITY it costs $7\%$ more at equal accuracy. Its per-benchmark average cost (as in \figref{fig:avg_bars}) is $0.77\times$ BL's without BoolQ and SIQA, whose single-call BL puts their ratios at $32\times$ and $4.5\times$.

\paragraph{Cross-model transfer.}
A strategy is most useful if it captures structure of the \emph{task} rather than idiosyncrasies of the executor it was searched on. We re-run \method{}-discovered strategies, without any modification to the blueprint code or its prompts, on two model families the strategist never saw during search: OpenAI GPT-5 (nano, mini, full) and Google Gemini-2.5 (flash-lite, flash). 
On GPT-5 the strategies win on 7 of 10 benchmarks and lose on 3 (ASDiv, HellaSwag, MinervaM: within 2 points of the baseline but costlier); on Gemini they win on 8 of 10, with SQuAD2 and MinervaM mixed (cheaper but less accurate). Aggregate cost in USD drops by $35\%$ and $40\%$. Part of the GPT-5 margin comes from baseline workflows that break on the new backend (\texttt{self\_refine} and \texttt{multi\_agent\_debate} return malformed answers there) while the discovered strategies run unchanged. That a Qwen3-discovered blueprint carries over at all shows that \method{} captures the structure of the task, not executor-specific tricks.

\begin{figure}[!t]
\centering
\includegraphics[width=0.79\linewidth]{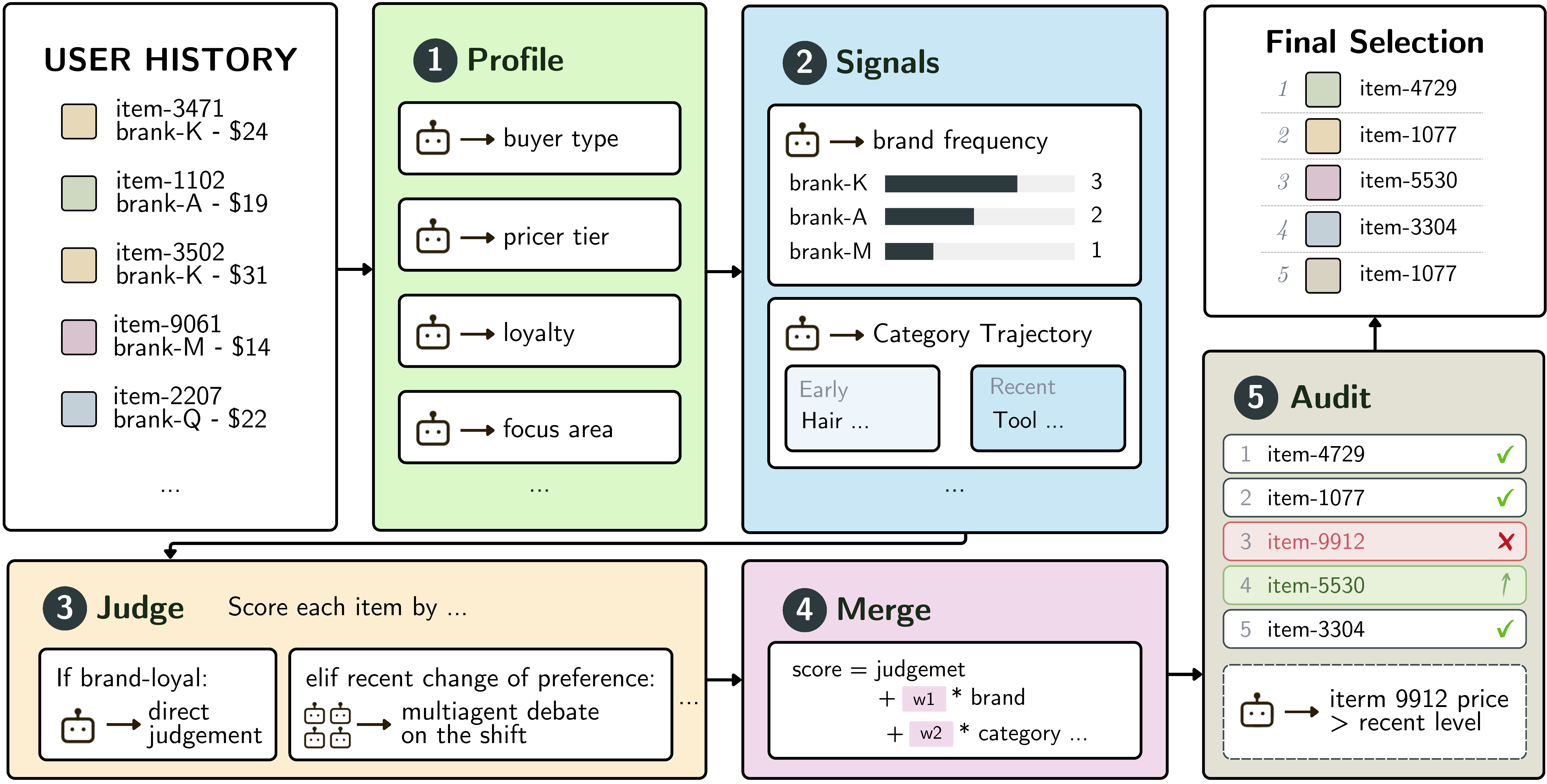}
\caption{\textbf{An example reasoning strategy discovered by \method{} for the Amazon-Rec task.} The strategy decomposes the task into a profile, signal extraction, conditional judge, merge, and audit stage. The structure and branch conditions were generated by the strategist, not by us.}
\label{fig:workflow_example}
\end{figure}

\section{Analysis}
\label{sec:analysis}

\paragraph{Anatomy of a discovered reasoning strategy.}
\figref{fig:workflow_example} shows one strategy (accuracy driven) designed for the Amazon-Rec task: it builds a user profile with cheap calls, extracts brand and category signals, and conditionally routes to a multi-agent debate only for users with shifting preferences, before merging and auditing the top picks. The branching is cheap deterministic logic that decides which inputs deserve the expensive call, so the strategy spends its capacity where it actually changes the answer rather than uniformly across all inputs. The conditional routing, the merge weighting, and the audit step all emerged from iterative proposals against measured accuracy. Because the result is plain Python, a user can read and edit it directly, and the call is structural, which is hard to do with prompt-bag methods or training-based alternatives.

\begin{figure}[!t]
\centering
\makebox[\linewidth][l]{\includegraphics[width=0.98\linewidth]{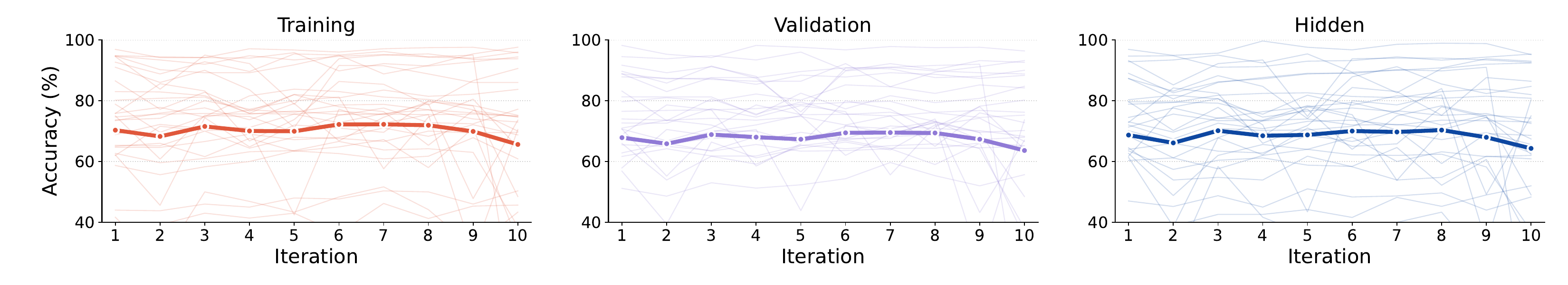}}
\vspace{-.65cm}
\caption{\textbf{Accuracy across data splits over the 10-iteration search loop.} Each point is the mean accuracy of an iteration's five candidates on a split. Bold curves are the mean across the 30 main benchmarks; faint background curves are individual benchmarks. The training--hidden gap stays between $1.2$ and $2.5$ points, not widening over iterations.}
\label{fig:overfitting}
\vspace{-.55cm}
\end{figure}

\paragraph{Overfitting across data splits.}
The strategist sees training traces and validation scores at every iteration and could fit noise in them; if it did, the training curve would pull away from the hidden curve. \figref{fig:overfitting} plots, per iteration, the mean accuracy of that iteration's five candidates on each split, averaged over the 30 main benchmarks. The curves move together and the training--hidden gap stays between $1.2$ and $2.5$ points, so the proposals do not drift away from data the strategist never sees. Selection is a different matter: the final strategy is chosen for being best on the hidden split, and part of that is luck. Across the 30 benchmarks the pick is $4.3$ points above BL on hidden but $0.7$ on test (BL itself moves by $0.5$). The test split, on which nothing is chosen, is the only unbiased estimate, and it is what every table reports.

\section{Ablations and Comparisons}
\label{sec:ablations}

\vspace{-.1cm}

To probe which components of \method{} are responsible for the gains in Section~\ref{sec:results}, we run two ablations on a fixed 10-benchmark set. Neither ablation falls back to BL, for testing purposes.
They are drawn from the 30 main ones, disjoint from transfer and one-shot sets.

\begin{table}[!b]
\centering
\footnotesize
\setlength{\tabcolsep}{3pt}
\begin{tabular}{lccc@{\hspace{8pt}}ccccc}
\toprule
& \multicolumn{3}{c}{\method{} variants} & \multicolumn{5}{c}{Other frameworks} \\
\cmidrule(lr){2-4} \cmidrule(lr){5-9}
Benchmark & full & no library & zero-shot & AFlow & ADAS & EvoFlow & StrategyLLM & AgentSquare \\
\midrule
BBH            & \textbf{85.0} & 31.0 & 27.5 & 38.0 & 34.5 & 34.5 & 35.5 & 27.5 \\
Gaokao-Math    & \textbf{94.5} & 65.5 & 81.5 & 84.0 & 82.0 & 86.5 & 89.5 & 82.5 \\
LSAT-LR        & \textbf{82.0} & 81.0 & 71.5 & 81.0 & 74.5 & 75.5 & 77.0 & 73.5 \\
LSAT-RC        & \textbf{81.5} & 75.0 & 38.0 & 74.5 & 75.5 & 75.0 & 74.5 & 74.0 \\
SAT-Math       & 98.2 & \textbf{99.0} & failed & \textbf{99.0} & 96.9 & 96.9 & 97.9 & 97.9 \\
MMLU-STEM-Hard & \textbf{88.0} & 83.5 & 86.5 & 83.0 & 82.5 & 87.5 & 86.0 & 86.5 \\
MMLU           & \textbf{79.5} & 78.0 & failed & 71.0 & 74.5 & 72.0 & 73.0 & 71.5 \\
MuSiQue        & \textbf{42.0} & 23.0 & 16.0 & 11.5 & 8.5 & 20.5 & 17.5 & 9.5 \\
CommonsenseQA  & 76.5 & \textbf{78.0} & 72.5 & 74.5 & 77.0 & 74.0 & 75.0 & 74.5 \\
AIME~2024      & \textbf{60.0} & 40.0 & failed & 53.3 & 53.3 & 20.0 & 46.7 & 53.3 \\
\midrule
\textbf{Average Acc}        & \textbf{78.7} & 65.4 & 39.4 & 67.0 & 65.9 & 64.2 & 67.3 & 65.1 \\
\bottomrule
\end{tabular}
\vspace{-0.1em}
\caption{\textbf{Ablations and cross-framework comparison on 10 benchmarks.} Test accuracy (\%); best per row bolded. External frameworks share the same meta-LLM and executor; ``failed'' is runtime crash.}
\vspace{-1em}
\label{tab:ablations}
\end{table}

\vspace{-.1cm}

\paragraph{Zero-shot selection.}
We disable the library and cap the loop at one iteration. The strategist sees only the task description and sample questions. It averages $39.4\%$ with three crashed runs counted as zero ($56.2\%$ over the seven that ran), $39.3$ points below ($78.7\%$).

\vspace{-.1cm}

\paragraph{Iteration without the library.}
We restore the 10-iteration loop but disable the library to isolate its contribution. This averages $65.4\%$ at $41.4k$ cost (\tabref{tab:ablations}), still $13.3$ points below ($78.7\%$). The main observation is that, without inspirations, it goes on to a narrow repetition.

\vspace{-.1cm}

\paragraph{Cross-framework comparison.}
We compare against five published frameworks (AFlow~\citep{zhang2024aflow}, ADAS~\citep{hu2024adas}, EvoFlow~\citep{zhang2025evoflow}, StrategyLLM~\citep{gao2023strategyllm}, AgentSquare~\citep{shang2024agentsquare}), all sharing the same meta-LLM and executor. \method{} wins on $8$ of $10$ benchmarks at $78.7\%$ mean accuracy. Many of these strategies fit the few development questions and fail on the test set.

\section{Related Work}
\label{sec:related}

\paragraph{LLM agents.}
LLM agents span chain-of-thought (CoT) and self-consistency reasoning~\citep{wei2022cot, wang2023selfconsistency}, deliberate search~\citep{yao2023tot}, tool-use loops~\citep{yao2023react, schick2023toolformer}, verbal self-reflection~\citep{shinn2023reflexion}, multi-agent conversation~\citep{wu2023autogen}, and formal representations from state machines to declarative compilation to optimizable graphs~\citep{wu2024stateflow, khattab2024dspy, zhuge2024gptswarm, survey2025acg}. The strategy layer (how models are called, in what order, and with what verification) has become the central design bottleneck; \method{} automates its design. (An extended discussion of related work is in Appendix~\ref{app:related_extended}.)

\paragraph{Automated design of agentic systems.}
Structure-search methods write strategies as raw code~\citep{hu2024adas} or compose them from fixed components or predefined modules~\citep{zhang2024aflow, shang2024agentsquare, zhang2025evoflow}; the first yields isolated programs, the second bounds what can be discovered. Optimization methods refine one strategy per task, by reinforcement learning~\citep{autoflow2025, zhang2026flowsteer, flowreasoner2025} or by LLM-driven evolution and refinement~\citep{gao2023strategyllm, sew2025}, and start over on the next task. Routing methods select among existing strategies without building them~\citep{arc2025, yue2025masrouter, bingreeky2025maas}. \method{} instead allows freeform creation via the Strategy Blueprint, needs no training, builds strategies itself, and routes across model tiers in each strategy.

\paragraph{Cross-task transfer and experience accumulation.}
Cross-task transfer has been studied through meta-learning~\citep{finn2017maml, hospedales2022metalearning}, neural architecture search~\citep{elsken2019nas}, and prompt optimization~\citep{yang2024opro}; in the LLM-agent setting, ExpeL~\citep{zhao2024expel} and Voyager~\citep{wang2023voyager} accumulate experience or skills, and lifelong-agent learning has been formalized as a direction~\citep{lifelong2025roadmap}. Existing automated design methods~\citep{hu2024adas, zhang2024aflow} still largely optimize per task. \method{} maintains a shared blueprint library that grows with every task, providing both exploitation (relevant components) and exploration (diverse inspiration); to our knowledge, the first automated strategy-design system that accumulates knowledge from one task to the next.

\section{Conclusion}
\label{sec:conclusion}

We presented \method{}, a meta-agent that designs inference strategies as freeform executable programs on a layered framework, the Strategy Blueprint, and grows a shared Blueprint Library across tasks. On 30 benchmarks, \method{}-designed strategies match or beat most strongest baselines on accuracy while running at substantially lower cost, and the discovered strategies transfer to backends the strategist never searched on. We view freeform strategy design over an accumulating library as a step toward inference systems that get better the more they are used.

\subsection*{Acknowledgments}
This work is supported by the computational resources generously provided by Google’s TPU Research Cloud program.  
This work was also performed using Princeton Research Computing resources, a consortium led by the Princeton Institute for Computational Science and Engineering (PICSciE) and Research Computing at Princeton University. 
We also gratefully acknowledge the use of the Neuronic GPU computing cluster maintained by the Department of Computer Science at Princeton University. 

\newpage


\bibliographystyle{plainnat}
\bibliography{main}

\newpage

\appendix

{%
\hypersetup{linkcolor=black}%
\section*{Appendix Contents}
\setlength{\parskip}{0pt}\setlength{\parindent}{0pt}%
\newcommand{\tocline}[2]{\hyperref[#2]{#1}\nobreak\leaders\hbox{\kern1pt.\kern1pt}\hfill\nobreak\pageref{#2}\par}
\newcommand{\tocsec}[1]{\vspace{3pt}\textbf{#1}\par}
\newcommand{\tocsub}[3]{\hspace*{1.5em}\tocline{#1}{#2}}

\tocsec{A.\enspace Method and Implementation Details}
\tocsub{A.1\enspace Pseudocode for the \method{} Design Loop}{app:algo}{}
\tocsub{A.2\enspace Benchmark and Baseline Settings}{app:bench_baseline}{}
\tocsub{A.3\enspace Data Split and Scaling}{app:data_split}{}
\tocsub{A.4\enspace Search, Library, and Models}{app:search_library}{}
\tocsub{A.5\enspace Strategist Prompt Template}{app:strategist_prompt}{}
\tocsub{A.6\enspace Cost-Weight Derivation}{app:cost_weights}{}
\tocsub{A.7\enspace Cross-Model Transfer Setup}{app:transfer_setup}{}

\tocsec{B.\enspace Examples and Case Studies}
\tocsub{B.1\enspace Sample Discovered Reasoning Strategy}{app:sample_workflow}{}
\tocsub{B.2\enspace Additional Sample Reasoning Strategies}{app:more_workflows}{}
\tocsub{B.3\enspace Sample Measurement Card}{app:sample_card}{}
\tocsub{B.4\enspace Search Queries and Experience Notes}{app:search_examples}{}
\tocsub{B.5\enspace Per-Iteration Progression Example}{app:iteration_log}{}

\tocsec{C.\enspace Per-Benchmark Result Tables}
\tocsub{C.1\enspace Per-Benchmark Main Results}{app:main_table}{}
\tocsub{C.2\enspace Per-Benchmark One-Shot Library Retrieval}{app:oneshot_full}{}
\tocsub{C.3\enspace Per-Benchmark Cross-Model Transfer}{app:transfer_full}{}


\vspace{3pt}\noindent\tocline{\textbf{D.\enspace Extended Related Work}}{app:related_extended}
}
\vspace{0.5em}

\section{Method and Implementation Details}

\subsection{Pseudocode for the \method{} Design Loop}
\label{app:algo}

Algorithm~\ref{alg:loop} formalizes the four-stage iterative pipeline described in Section~\ref{sec:meta}. Inputs are a task description $\mathcal{T}$, a development set $D$ split into feedback/validation/hidden subsets, the current shared blueprint library $\mathcal{L}$, and the number of iterations $N$. Each iteration runs four steps: (i)~the strategist queries $\mathcal{L}$ for the most relevant blueprints and a set of adjacent-domain inspirations; (ii)~the strategist proposes $K$ candidate strategies, balanced across the kinds \emph{new}/\emph{composed}/\emph{reuse} and the targets \emph{accuracy}/\emph{efficiency}/\emph{balanced}; (iii)~each candidate is executed on $D$ and scored on accuracy and cost; (iv)~the strategist writes experience notes summarizing what worked and what failed, and the candidates plus their measurement cards are written back to $\mathcal{L}$ with structural deduplication. The reference baseline $b$ (the baseline cell most accurate on the training split, run with its exact code) is evaluated alongside the candidates in the first iteration. After $N$ iterations, every candidate is compared with $b$ question by question on the validation and hidden splits; among the candidates that beat $b$ under the 2\% rule on both (better by more than 2 points in accuracy or 2\% in cost, worse on neither), the one with the largest hidden-set accuracy gain is returned as $s^*$, and $b$ itself if none qualifies. The updated library $\mathcal{L}'$ is carried forward for future tasks.

\begin{algorithm}[h!]
\caption{The \method{} design loop}
\label{alg:loop}
\fontfamily{cmss}\selectfont\small
\begin{algorithmic}[1]
\REQUIRE Task description $\mathcal{T}$, development set $D$, library $\textcolor{alglib}{\mathcal{L}}$, iterations $N$, reference baseline $b$
\ENSURE Winning strategy $s^*$, updated library $\textcolor{alglib}{\mathcal{L}'}$
\FOR{$i = 1$ \TO $N$}
    \STATE \textcolor{algcmd}{Strategist} queries $\textcolor{alglib}{\mathcal{L}}$ for relevant blueprints and diverse inspiration
    \STATE \textcolor{algcmd}{Strategist} proposes $K$ candidate strategies $\{s_1, \dots, s_K\}$
    \FOR{each candidate $s_j$}
        \STATE \textcolor{algeval}{Evaluate} $s_j$ on $D$: measure \textcolor{algeval}{accuracy} and \textcolor{algeval}{cost}
    \ENDFOR
    \STATE \textcolor{algcmd}{Strategist} analyzes \textcolor{algeval}{results} and writes experience notes
    \STATE Add each $s_j$, its measurement card, and experience notes to $\textcolor{alglib}{\mathcal{L}}$ (with deduplication)
\ENDFOR
\STATE $W \leftarrow$ candidates that beat $b$ under the 2\% rule on both validation and hidden
\STATE $s^* \leftarrow \arg\max_{s \in W} \textcolor{algeval}{\Delta\text{accuracy}}_{\text{hidden}}(s, b)$, or $b$ if $W = \emptyset$
\RETURN $s^*$, $\textcolor{alglib}{\mathcal{L}}$
\end{algorithmic}
\end{algorithm}

\subsection{Benchmark and Baseline Settings}
\label{app:bench_baseline}

\paragraph{Scoring mechanism.}
All strategies and baselines run as 0-shot generation: the executor LLM produces a free-form response and an answer extractor parses the final answer from the trace. The extractor is benchmark-aware (multiple-choice letter, integer, LaTeX numeric, span, or yes/no) and is the same across \method{}, all baselines, and all five comparison frameworks, so reported accuracy differences reflect strategy choices rather than scoring asymmetries.

\paragraph{Benchmarks.}
We evaluate on 40 publicly released benchmarks. From these, $30$ are randomly selected as the main benchmark set used for the iterative search and the remaining $10$ are held out exclusively for the one-shot retrieval experiment.

\begin{itemize}[leftmargin=*, topsep=2pt, itemsep=1pt]
    \item \textbf{Mathematical reasoning:} GSM8K~\citep{cobbe2021gsm8k} (grade-school word problems, numeric), MATH500~\\ 
    \citep{hendrycks2021math} (competition math, LaTeX/numeric), AQuA-RAT~\citep{ling2017aqua} (algebra, 5 choices), AIME~2024~\citep{aime2024} (American Invitational Math, 30 problems, integer 0--999), ASDiv~\citep{miao2020asdiv} (diverse word problems, numeric), SVAMP~\citep{patel2021svamp} (perturbed word problems, numeric), MinervaMath~\citep{lewkowycz2022minerva} (graduate-level quantitative, LaTeX/numeric), Gaokao-Math-QA~\citep{zhong2023agieval} (Chinese college-entrance MCQ, 4 choices).
    \item \textbf{Code generation:} MBPP+~\citep{liu2023evalplus} (Python function synthesis, pass@1 on extra-test variant of MBPP).
    \item \textbf{Commonsense reasoning:} HellaSwag~\citep{zellers2019hellaswag} (sentence completion, 4 choices), PIQA~\citep{bisk2020piqa} (physical commonsense, 2 choices), Social IQa~\citep{sap2019siqa} (social commonsense, 3 choices), WinoGrande~\citep{sakaguchi2020winogrande} (coreference, 2 choices), CommonsenseQA~\citep{talmor2019csqa} (general commonsense, 5 choices), COPA~\citep{roemmele2011copa} (causal alternatives, 2 choices), OpenBookQA~\citep{mihaylov2018openbookqa} (elementary science, 4 choices), SciQ~\citep{welbl2017sciq} (crowdsourced science, 4 choices), StrategyQA~\citep{geva2021strategyqa} (implicit-reasoning yes/no), BoolQ~\citep{clark2019boolq} (passage-grounded yes/no).
    \item \textbf{Knowledge and QA:} MMLU~\citep{hendrycks2021mmlu} (57 subjects, 4 choices), MMLU-STEM-Hard (filtered hard STEM subset, 4 choices), TriviaQA~\citep{joshi2017triviaqa} (closed-book trivia, EM), NQ-Open~\citep{kwiatkowski2019nq} (closed-book Wikipedia QA, EM), TruthfulQA~\citep{lin2022truthfulqa} (misconception-resistant MC1), SQuAD~v2~\citep{rajpurkar2018squad2} (extractive with unanswerables, F1).
    \item \textbf{Multi-hop QA:} HotpotQA~\citep{yang2018hotpotqa} (2-hop with passages, EM), 2WikiMultihopQA~\citep{ho2020twowiki} (Wikipedia multi-hop, EM), MuSiQue~\citep{trivedi2022musique} (2--4-hop composition, EM), DROP~\citep{dua2019drop} (discrete reasoning over paragraphs, EM), QuALITY~\citep{pang2022quality} (long-passage MCQ on 5\,k-token articles, 4 choices).
    \item \textbf{Logical and analytical reasoning:} BBH~\citep{suzgun2023bbh} (23 BIG-Bench Hard tasks, mixed format), ANLI-R3~\citep{nie2020anli} (round-3 adversarial NLI, 3 choices), ReClor~\citep{yu2020reclor} (logical reading comprehension, 4 choices), AGIEval English MCQ~\citep{zhong2023agieval}: SAT-Math, SAT-EN, LSAT-AR, LSAT-LR, LSAT-RC (4--5 choices each), MedQA~\citep{jin2021medqa} (USMLE-style, 4 choices).
    \item \textbf{Recommendation:} Amazon-Rec~\citep{liu2025onerecthink} (rank top item from a candidate pool, hit-rate).
\end{itemize}

\paragraph{Baselines.}
We compare against $18$ standard baselines covering both single-call prompting and full agentic systems. Each baseline is evaluated on every one of the six tier/mode configurations, giving an $18 \times 6 = 108$-cell baseline matrix per benchmark. The matrix is measured on both the training split and the test set; the cell most accurate on the training split is BL, and its test result is reported.

\begin{itemize}[leftmargin=*, topsep=2pt, itemsep=2pt]
    \item \textbf{direct:} a single LLM call that returns the answer with no intermediate reasoning. The minimum-cost baseline; serves as the floor for comparisons on easy benchmarks.
    \item \textbf{chain-of-thought (CoT)}~\citep{wei2022cot}: a single call prompted with ``Let's think step by step'', encouraging the model to emit intermediate reasoning before the final answer.
    \item \textbf{self-consistency}~\citep{wang2023selfconsistency}: sample $3$ CoT trajectories at temperature $0.7$ and return the majority-voted final answer; trades cost for variance reduction.
    \item \textbf{direct-thinking:} a single call in \texttt{thinking} mode with no extra prompting, isolating the contribution of the model's hidden reasoning trace from any prompting trick.
    \item \textbf{tree-of-thoughts (ToT)}~\citep{yao2023tot}: maintain a tree of partial reasoning states, expand each by sampling a few continuations, score with the model itself, and search BFS-style for a high-scoring leaf.
    \item \textbf{least-to-most}~\citep{zhou2023leasttomost}: a first call decomposes the question into sub-problems; subsequent calls solve them in order, each conditioned on prior answers.
    \item \textbf{chain-of-verification (CoVe)}~\citep{dhuliawala2023cove}: produce a draft answer, generate verification questions about it, answer those independently, and revise if any verification disagrees.
    \item \textbf{self-discover}~\citep{zhou2024selfdiscover}: the model first selects and composes a custom reasoning structure (decomposition, verification, etc.) from a fixed pool, then applies that structure to the input.
    \item \textbf{self-refine}~\citep{madaan2023selfrefine}: a feedback-and-revise loop in which the model critiques its own draft and rewrites it; we cap the loop at three rounds.
    \item \textbf{vanilla-agent:} an OpenHands ReAct loop with the \texttt{TerminalTool} and \texttt{FileEditorTool} only, no extra prompt scaffolding.
    \item \textbf{agent-code:} the same agent but the prompt instructs the model to write and execute Python via the terminal tool, so numeric/code-heavy steps are grounded in actual computation.
    \item \textbf{multi-agent-debate}~\citep{du2024debate}: $3$ agents propose answers, then in $2$ subsequent rounds each agent sees the others' answers and revises; the final majority vote is returned.
    \item \textbf{MetaGPT}~\citep{hong2023metagpt}: a role-based pipeline (planner, executor, reviewer) that hands a structured artifact between roles; we use the open-source role prompts unchanged.
    \item \textbf{AutoGen}~\citep{wu2023autogen}: a two-agent conversable setup (assistant + user-proxy with tool-use) that converses until the assistant emits a termination signal.
    \item \textbf{chain-of-agents (CoA)}~\citep{zhang2024chainofagents}: a chain of worker agents each summarises a slice of the input and passes a hand-off message to the next; a final manager agent answers from the chain.
    \item \textbf{reflexion}~\citep{shinn2023reflexion}: an outer loop that, on failure, asks the model to write a verbal self-reflection and retries with the reflection appended to the next prompt.
    \item \textbf{monte-carlo-vote / shortlist-vote:} two self-consistency variants---the first samples $5$ traces and returns the strict majority (resampling on a tie), the second pre-filters to a top-$10$ shortlist before a $3$-way vote---used to test whether voting beyond the standard $3$ samples helps.
\end{itemize}

All baselines run on the same Qwen3 executors and the same OpenHands tool sandbox (\texttt{TerminalTool} + \texttt{FileEditorTool} only) with no internet access, identical to \method{} strategies. Each baseline is evaluated independently at every one of the six tier/mode configurations, but baselines do not internally route across tiers within a single run; the ability to mix tiers within one strategy is part of the design space \method{} searches over, and BL is the single-config cell most accurate on the training split.

\subsection{Data Split and Scaling}
\label{app:data_split}

Every benchmark is partitioned into four disjoint subsets: \emph{training}, \emph{validation}, \emph{hidden}, and \emph{test}. The training set is what the strategist sees in full, including per-sample reasoning traces, when proposing the next iteration's slate. The validation set is also evaluated each iteration but only aggregate scores are visible, mirroring standard train/val practice. The validation and hidden sets are used to pick the final strategy across all $N$ iterations (Algorithm~\ref{alg:loop}); the hidden set is never shown to the strategist; in particular, experience notes written back to the library are derived from training-set traces only, so hidden-set signal cannot leak across iterations or to future tasks. The test set is held out from the search loop entirely and is used only for the final number reported in the main text.

\paragraph{Sizes and caps.}
The development rows of each benchmark come from a separate development pool (300 rows where available), drawn deterministically with a fixed seed and split into training, validation, and hidden sets of $100$ rows each; benchmarks with smaller pools scale all three down proportionally, and AIME~2024 and Amazon-Rec, whose questions are about $10\times$ longer to solve, use $50$ rows per split. Where a benchmark's own pool was too small, it was extended: COPA and SVAMP with rows from their official training data, and AIME with the 299 problems of AIME 2014--2023. 
The test set is around the first $200$ rows of each benchmark's test file, disjoint from all development rows and never seen during search. Six benchmarks have fewer questions due to limit data. 
Splits are deterministic, so reruns and ablations operate identically.

\subsection{Search, Library, and Models}
\label{app:search_library}

\paragraph{Search loop.}
Each iteration proposes $K=5$ candidate strategies; the search runs for $N=10$ iterations per benchmark. Every iteration must include at least one candidate of each kind (\emph{new}, \emph{composed}, \emph{reuse}) and of each target (\emph{accuracy}: beat BL's accuracy within its cost; \emph{efficiency}: hold its accuracy at lower cost; \emph{balanced}); slates that violate this rule, deduplicate a recent candidate, or omit required JSON fields are rejected and the strategist is re-prompted. The strategist is shown the baseline matrix measured on the training split (every baseline at all six configurations, with BL highlighted) and the 2\% rule by which candidates are judged; it keeps one session across iterations, so later iterations send only the new results. Each LLM call from a strategy uses \texttt{max\_tokens=16384}. A full search evaluates about 60 candidates per benchmark ($K{=}5$ per iteration plus the reference and repairs), each on all three development splits, roughly $20{,}000$ candidate-sample executions for a benchmark with $100$ rows per split, plus about $135$k prompt tokens and $45$k response tokens for the strategist (about $10$k prompt tokens per iteration), so the search dominates the per-task cost and amortises across future tasks via library reuse (Section~\ref{sec:results}).

\paragraph{Deduplication.}
Two blueprints are considered duplicates if their \texttt{execute} function bodies hash to the same value after task-specific string literals (prompts, format instructions) are elided; this is computed over the Python AST so that whitespace and comment changes do not produce spurious distinctions. Duplicate proposals are rejected at submission time and not added to the library.

\paragraph{Strategist proposal format.}
Each iteration the strategist returns Python code (the candidate strategies) followed by a JSON block describing the slate. The JSON has six required fields per candidate (\texttt{name}, \texttt{kind}, \texttt{target}, \texttt{tier\_mode\_anchors}, \texttt{why}, \texttt{builds\_on}) and an optional \texttt{prompts} field that replaces a blueprint's prompt texts for this task, plus three search-side fields (\texttt{search\_queries}, \texttt{read\_cards}, \texttt{experience}). A non-fatal AST audit checks that \texttt{builds\_on} matches the imports in the code block; warnings are fed into the next iteration's prompt rather than killing the run.

\paragraph{Library retrieval.}
At each iteration the strategist issues two textual queries against the library: a relevance query summarising what the current task is asking for, and an inspiration query phrased to deliberately drift from the task description (e.g., asking for ``multi-path verification on long-context QA'' when the task itself is short-form math). Each query is scored by a hybrid function $s = \alpha\,s_{\text{tfidf}} + (1-\alpha)\,s_{\text{emb}}$ over every blueprint's index entry, where $s_{\text{tfidf}}$ is computed over the blueprint name, one-line description, when-to-use field, and trait tags, and $s_{\text{emb}}$ is the cosine similarity between the query and a sentence-transformer (\texttt{all-MiniLM-L6-v2}) embedding of the blueprint's measurement card and experience notes; we use $\alpha=0.4$ throughout. Each query returns its top-$5$ blueprints, deduplicated against blueprints already returned by the relevance query so the inspiration slate cannot overlap. The strategist sees a total of $10$ blueprints per iteration (5 relevance hits with their measurement cards rendered in full, 5 adjacent inspirations with only the one-line description shown) and is asked, in the prompt, to use the relevance hits as composition or reuse candidates and the inspiration hits as material for new blueprints.

\paragraph{Library size.}
After the 30 main searches, the library contained $1{,}865$ blueprints (deduplicated by structural hash) with measurement cards (cards are written only after a blueprint has been evaluated at least once). Of these, the strategist chose only those with relevance scores above a fixed threshold during retrieval; on average, $5$ relevance hits and $5$ adjacent inspirations were returned per query.

\paragraph{Models.}
The executor LLMs are Qwen3-0.6B (\texttt{fast} tier), Qwen3-1.7B (\texttt{balanced}), and Qwen3-4B (\texttt{complex}), each available in \texttt{nothink} and \texttt{thinking} modes for a total of six configurations; cross-model transfer (Section~\ref{sec:results}, Appendix~\ref{app:transfer_setup}) maps each tier to GPT-5 (nano, mini, full) or Gemini-2.5 (flash-lite, flash). The strategist agent is GPT-6 (\texttt{gpt-6-sol}), accessed through the Codex CLI. All five comparison frameworks in the cross-framework table use the same strategist and executor as \method{} for fairness.

\paragraph{Infrastructure.}
The Qwen3 executors are served via vLLM on a local cluster of $8\times$ NVIDIA H200 GPUs (144\,GB each). Agent-tier strategies execute through OpenHands with \texttt{TerminalTool} and \texttt{FileEditorTool} only; we cap subprocess concurrency at 48 to avoid timeouts. All hyperparameters above are fixed across the 30 main benchmarks, the 10 one-shot held-out benchmarks, and the cross-model transfer experiments.

\subsection{Strategist Prompt Template}
\label{app:strategist_prompt}

The strategist receives the full prompt in the first iteration and keeps one session; later iterations send only what is new (results, library hits, the updated history). Listing~\ref{lst:strategist_prompt} shows the abridged template.

\begin{lstlisting}[style=promptstyle, basicstyle=\sffamily\scriptsize, captionpos=b, caption={Abridged strategist prompt template (GPT-6).}, label={lst:strategist_prompt}]
[SYSTEM]
You are the Strategist meta-agent. Design 5 candidate strategies for the task.
Goal: beat the reference baseline (BL) on this benchmark. BL's accuracy is the
floor and its cost per sample the budget; within 2 points of accuracy or 2% of
cost is a tie. A win is lower cost at no worse accuracy, or higher accuracy at
no higher cost.
A strategy is an executable Python program:
    async def execute(llm, task: str) -> str        (one tier/mode), or
    async def execute(llms, task: str) -> str       (routes across tiers)
Models: {fast,balanced,complex} x {nothink,thinking}; library blueprints are
queryable by name, traits and measurement cards.
Every slate has >=1 of each kind (new / composed / reuse) and of each target
(accuracy / efficiency / balanced).
[USER]
Task: {task_description}; sample training questions with answers
Baselines on the training split: {18 x 6 table, BL highlighted}
Library hits:
  Relevant: {top_5_relevant_blueprints_with_cards}
  Adjacent inspiration: {top_5_adjacent_blueprints}
Results so far: {each candidate vs BL on the training split: accuracy, cost
  ratio, questions fixed/broken}
Produce (1) Python code for 5 candidates then (2) a JSON block with fields:
candidates[5]={name,kind,target,tier_mode_anchors,why,builds_on,prompts},
search_queries, read_cards, experience[]={workflow,observation,recommendation}.
\end{lstlisting}

The \texttt{experience} field is the strategist's own writeback to the library: every blueprint it referenced or used must come back with one observation (numerical, e.g., ``card shows +12\% on math/fast/nothink'') and one recommendation (concrete, e.g., ``use as cheap lift on short-math fast-tier; skip on commonsense''). These notes feed forward into future strategists' card lookups.

\subsection{Cost-Weight Derivation}
\label{app:cost_weights}

We use two cost units in the paper. For the Qwen3 main experiments, where we control the inference stack, we report \emph{weighted tokens}: per-tier weights proportional to parameter count, justified below. For the cross-model transfer experiments on GPT-5 and Gemini-2.5, we report \emph{USD} computed at the official list prices published on the OpenAI\footnote{\url{https://openai.com/api/pricing/}} and Google\footnote{\url{https://ai.google.dev/gemini-api/docs/pricing}} pricing pages, applied identically to baseline and \method{} runs so that any difference reflects strategy choices rather than assumptions.

\paragraph{Qwen3 weights.}
For a dense decoder-only transformer, per-token forward FLOPs scale as ${\approx}2N$ where $N$ is the parameter count~\citep{kaplan2020scaling}. The Qwen3 tiers used in the main experiment have $0.6$B, $1.7$B, and $4$B parameters, giving weight ratios of $1.00 : 2.83 : 6.67$ (\tabref{tab:cost_weights}). Input tokens count at the same per-token weight, and tool execution for agent strategies (a sandboxed terminal) is not charged. Thinking-mode tokens are counted along with output tokens; their accuracy lift is paid for in the same currency.

\begin{table}[!htbp]
\centering
\footnotesize
\setlength{\tabcolsep}{6pt}
\begin{tabular}{lccc}
\toprule
Tier & Model & Parameters & Cost weight \\
\midrule
fast     & Qwen3-0.6B          & $0.60$B & $1.00\times$ \\
balanced & Qwen3-1.7B          & $1.70$B & $2.83\times$ \\
complex  & Qwen3-4B            & $4.00$B & $6.67\times$ \\
\bottomrule
\end{tabular}
\caption{\textbf{Tier cost weights used throughout the paper.} Per-token weights are proportional to parameter counts, matching the leading $2N$ FLOPs/token term for dense transformers~\citep{kaplan2020scaling}.}
\label{tab:cost_weights}
\end{table}

\subsection{Cross-Model Transfer Setup}
\label{app:transfer_setup}

Cross-model transfer (Section~\ref{sec:results}) runs the same \method{}-discovered strategies on backends never used during search. The tier-mode mapping is fixed per backend; nothing is re-tuned. \tabref{tab:transfer_tiers} reports the mapping.

\begin{table}[!htbp]
\centering
\footnotesize
\setlength{\tabcolsep}{6pt}
\begin{tabular}{lll}
\toprule
Tier/mode & GPT-5 backend & Gemini-2.5 backend \\
\midrule
fast/nothink     & gpt-5-nano (effort=minimal) & gemini-2.5-flash-lite \\
fast/thinking    & gpt-5-nano (effort=medium)  & gemini-2.5-flash-lite (thinking) \\
balanced/nothink & gpt-5-mini (effort=minimal) & gemini-2.5-flash \\
balanced/thinking& gpt-5-mini (effort=medium)  & gemini-2.5-flash (thinking) \\
complex/nothink  & gpt-5 (effort=minimal)      & gemini-2.5-flash \\
complex/thinking & gpt-5 (effort=medium)       & gemini-2.5-flash (thinking) \\
\bottomrule
\end{tabular}
\caption{\textbf{Tier-mode mapping for cross-model transfer.} GPT-5 uses three model sizes (nano/mini/full) crossed with two reasoning-effort levels (minimal/medium). Gemini-2.5 uses flash-lite for the fast tier and flash for the balanced and complex tiers; the API sets its own thinking budget, so the nothink and thinking modes issue the same call. Each strategy executes its own routing; only the tier-to-model mapping changes.}
\label{tab:transfer_tiers}
\end{table}

\section{Examples and Case Studies}

\subsection{Sample Discovered Reasoning Strategy}
\label{app:sample_workflow}

To make the structure of a discovered strategy concrete, we reproduce one (\texttt{abstraction\_ladder}) which the strategist proposed as a multi-path reasoning blueprint. The header declares its meta-information (when to use, traits, foundation); the \texttt{execute} function (abridged below) defines the call graph. The listings show the interface as implemented: \texttt{execute(llm, task, **prompts)}, where \texttt{llm} is one client or, for strategies that route across tiers, the dictionary of six clients (written \texttt{llms}), and prompt replacements arrive as keyword arguments.

\begin{lstlisting}[style=codestyle, language=Python, caption={The \texttt{abstraction\_ladder} blueprint: META declaration plus \texttt{execute} body. Multi-path reasoning at three abstraction levels; agreement between the two grounded levels is accepted, otherwise a fourth call adjudicates with all three visible. Prompt text is abridged into named helpers.}, label={lst:abstraction_ladder}, float=!ht]
META = {
    "name": "abstraction_ladder",
    "description": (
        "Solve at 3 abstraction levels (principle, specific, "
        "concrete). If all agree, high confidence. If "
        "disagreement, prefer the concrete-example answer as most grounded."
    ),
    "when_to_use": (
        "Multi-step reasoning or math problems where abstract "
        "reasoning might miss details but pure computation might "
        "miss the big picture."
    ),
    "tags": ["reasoning", "multi_path"],
    "uses": ["generate"],
    "traits": {
        "foundation": "generate",
        "reasoning": True,
        "self_improving": False,
        "multi_path": True,
        "tool_use": False,
        "decomposition": True,
        "multi_turn": True,
    },
}

async def execute(llm, task: str, **kwargs) -> str:
    system = f"{base_system(task)} {output_instruction(task)}"
    abstract  = await generate(llm, level1_prompt(task),  system)
    specific  = await generate(llm, level2_prompt(task, abstract), system)
    concrete  = await generate(llm, level3_prompt(task, specific), system)
    if extract_answer(task, specific) == extract_answer(task, concrete):
        return extract_answer(task, concrete)   # two grounded levels agree
    adjudicated = await generate(
        llm, adjudicate_prompt(task, abstract, specific, concrete), system)
    return extract_answer(task, adjudicated)
\end{lstlisting}

The strategist annotates every blueprint with \emph{traits} (above) so future searches can retrieve by structural similarity (\texttt{multi\_path}, \texttt{decomposition}, \ldots) rather than only by name. The \texttt{when\_to\_use} field is what subsequent strategists read to decide whether to compose on top of this blueprint.

\subsection{Additional Sample Reasoning Strategies}
\label{app:more_workflows}

To illustrate the diversity of structures the strategist discovers, we show three more final picks from the main run, each the strategy selected for its benchmark across all iterations (Algorithm~\ref{alg:loop}). The code is abridged from the library files: prompt text is replaced by named helpers, control flow is unchanged.

\paragraph{Logic (BBH): \texttt{high\_precision\_execution\_router}.}
A 0.6B router call decides whether a short program on the literal data in the task can compute and verify the answer. If so, the task goes to the library's tool-using agent blueprint at \texttt{complex/thinking}; otherwise to self-consistency at the same tier. The blueprint is composed from two library blueprints and adds only the gate. On BBH it scores 85.0\% against BL's 86.0\% at 0.55$\times$ the cost.

\begin{lstlisting}[style=codestyle, language=Python, caption={\texttt{high\_precision\_execution\_router}: a cheap router sends deterministic computations to the tool agent and everything else to self-consistency.}, label={lst:high_precision_execution_router}, float=!ht]
from strategist.workflows.baseline_agent_code import execute as agent_code
from strategist.workflows.baseline_self_consistency import execute as self_consistency

async def execute(llms, task):
    # EXECUTE if a program on the literal data can compute and verify the answer;
    # CONSENSUS for world knowledge, pragmatics, analogy, or when uncertain.
    route = await generate(llms["fast"]["nothink"], route_prompt(task))
    if route.strip().upper().startswith("EXECUTE"):
        result = await agent_code(llms["complex"]["thinking"], task)
    else:
        result = await self_consistency(llms["complex"]["thinking"], task)
    return extract_answer(task, result)
\end{lstlisting}

\paragraph{Multi-hop QA (MuSiQue): \texttt{strict\_roundtrip\_count\_multihop}.}
The strategy runs a library multi-hop blueprint and then guards one failure mode: a count answered as a digit when the passages spell it as a word. It asks a 1.7B model to spell the number, accepts the word only if it occurs in the context and an independent call decodes it back to the same number, and otherwise keeps the digit. On MuSiQue it scores 42.0\% against BL's 41.0\% at 0.93$\times$ the cost.

\begin{lstlisting}[style=codestyle, language=Python, caption={\texttt{strict\_roundtrip\_count\_multihop}: a library multi-hop blueprint followed by a round-trip check before a count is rewritten as a word.}, label={lst:strict_roundtrip_count_multihop}, float=!ht]
from strategist.workflows.question_bound_span_projection_multihop import projected_walk

async def execute(llms, task):
    answer = extract_answer(task, await projected_walk(llms, task)).strip(" .")
    context, _, question = task.rpartition("\nQuestion:")
    asks_for_count = ("how many" in question.casefold()
                      or "number of" in question.casefold())
    if asks_for_count and answer.isdecimal() and 0 <= int(answer) <= 12:
        word = (await generate(llms["balanced"]["nothink"],
                               spell_prompt(answer))).strip().lower()
        attested = (re.fullmatch(r"[a-z]{3,}", word)
                    and re.search(rf"\b{re.escape(word)}\b", context, re.I))
        if attested:
            decoded = await generate(llms["balanced"]["nothink"], decode_prompt(word))
            m = re.fullmatch(r"\s*(\d{1,2})\.?\s*", decoded)
            if m and m.group(1) == answer:
                answer = word
    return extract_answer(task, answer)
\end{lstlisting}

\paragraph{Math (GSM8K): \texttt{dual\_verbatim\_evidence\_lens\_math}.}
Two parallel \texttt{complex/nothink} calls copy verbatim spans from the problem, one for what is asked and one for counting scope and units; spans that do not occur in the problem are dropped. A single \texttt{complex/thinking} call then solves the original problem with the spans as an attention guide. On GSM8K it scores 92.0\% against BL's 94.0\% at 0.13$\times$ the cost.

\begin{lstlisting}[style=codestyle, language=Python, caption={\texttt{dual\_verbatim\_evidence\_lens\_math}: parallel verbatim evidence extraction, filtered against the source, then one solving call.}, label={lst:dual_verbatim_evidence_lens_math}, float=!ht]
async def execute(llms, task):
    target_text, scope_text = await asyncio.gather(
        generate(llms["complex"]["nothink"], copy_target_spans_prompt(task)),
        generate(llms["complex"]["nothink"], copy_scope_spans_prompt(task)))
    spans = [l.strip() for l in (target_text + "\n" + scope_text).splitlines()
             if l.strip() and l.strip() in task][:6]
    focus = "\n".join(spans) if spans else "No exact spans were selected."
    answer = await generate(llms["complex"]["thinking"],
                            solve_prompt(task, focus),
                            system=output_instruction(task))
    return extract_answer(task, answer)
\end{lstlisting}

These three strategies cover different control-flow patterns (route-then-delegate, compose-then-guard, parallel-extract-then-solve) yet share the same interface, so each can serve as a component of a later blueprint.

\subsection{Sample Measurement Card}
\label{app:sample_card}

After a strategy is run on a benchmark, the strategist writes a \emph{measurement card} summarizing what worked, what failed, and how to use the strategy next time. Cards are stored alongside their strategy and are returned as part of library retrieval. Below is a simplified, shortened excerpt of the card for \texttt{cite\_then\_extract}:

\begin{lstlisting}[style=cardstyle, caption={Excerpt of the measurement card for \texttt{cite\_then\_extract}, written by the strategist after evaluation. Cards are returned with their blueprint during retrieval and inform future strategists' design choices.}, label={lst:card_cite_then_extract}, float=!ht]
# cite_then_extract

## Summary
Two-step pipeline: first cite the supporting passage, then extract
the answer from the cited span. Reduces hallucination on multi-hop
QA where the document provides the evidence.

## Measured behavior
- 2WikiMQA: 74.5% acc (BL 66.5%, +12.0% rel) at 23.2k weighted
  tokens / sample.
- HotpotQA: 63.5% acc (BL 53.0%, +19.8% rel).
- MuSiQue:  24.5% acc (BL 19.5%, +25.6% rel).

## Strengths
- Stable on benchmarks where the evidence sits in a long context.
- Robust to distractor passages: citing first forces the model to
  commit to a span before answering.

## Weaknesses
- On short-context QA the citation step is wasted cost.
- If the cite step picks the wrong span, the extraction step
  inherits the error with no recovery.

## Notes from prior use
- 2026-04-22: Used as composition base for entity_focused_debate
  (HotpotQA). Composed candidate beat parent +5pp.

...
\end{lstlisting}

\subsection{Search Queries and Experience Notes}
\label{app:search_examples}

\tabref{tab:search_examples} samples actual \texttt{search\_queries} and \texttt{experience} notes the strategist wrote during search runs. These notes are the cross-task knowledge that future iterations and future tasks read; they are why the library compounds rather than starts from scratch each task.

\begin{table}[!htbp]
\centering
\footnotesize
\setlength{\tabcolsep}{4pt}
\begin{tabular}{p{0.3\linewidth}p{0.65\linewidth}}
\toprule
Field & Sample value \\
\midrule
\texttt{search\_queries}
  & ``cite-then-extract on long context multi-hop'' \\
  & ``self-consistency vote at fast tier with 3 samples'' \\
  & ``multi-agent debate but only on questionable cases'' \\
  & ``decomposition into sub-questions then aggregate'' \\
\midrule
\texttt{experience} (workflow, obs, rec)
  & \texttt{self\_consistency}: ``card shows +12pp on math/fast/nothink; blind matched at 11pp'' / ``use as cheap lift on short-math fast-tier; skip on commonsense.'' \\
  & \texttt{multi\_agent\_debate}: ``2-agent fails by mutual-error agreement on 30\% of NQ-Open misses'' / ``add a third tie-breaking agent or vote.'' \\
  & \texttt{cite\_then\_extract}: ``stable on long-context QA; wasted cost on short-context'' / ``compose only when context length $>$ 1024 tokens.'' \\
\bottomrule
\end{tabular}
\caption{\textbf{Sampled \texttt{search\_queries} and \texttt{experience} notes.} The strategist writes both at every iteration; queries are the input to library retrieval, and experience notes are the writeback that next-iteration strategists (or strategists on a different task) read in the card.}
\label{tab:search_examples}
\end{table}

\subsection{Per-Iteration Progression Example}
\label{app:iteration_log}

To make the iterative dynamics concrete, \tabref{tab:iteration_log} shows the per-iteration progression on AQuA-RAT. Each row reports the candidate of that iteration with the highest hidden-set accuracy; the final pick follows Algorithm~\ref{alg:loop}.

\begin{table}[!htbp]
\centering
\footnotesize
\setlength{\tabcolsep}{6pt}
\begin{tabular}{rllrr}
\toprule
Iter & Kind & Best candidate on hidden & Hidden Acc & Hidden Cost ($k$) \\
\midrule
1  & reference & \texttt{baseline\_cot (BL)} & 85.0 & 21.2 \\
2  & reuse     & \texttt{difficulty\_routed\_certificate\_mcq} & 89.0 & 18.5 \\
3  & reuse     & \texttt{difficulty\_routed\_certificate\_mcq} & 91.0 & 21.7 \\
4  & reuse     & \texttt{difficulty\_routed\_certificate\_mcq} & 92.0 & 20.9 \\
\textbf{5} & \textbf{composed} & \texttt{\textbf{intended\_operation\_commitment\_cot\_math}} & \textbf{90.0} & \textbf{16.2} \\
6  & new       & \texttt{rotated\_label\_invariance\_math} & 88.0 & 18.6 \\
7  & reuse     & \texttt{label\_permutation\_invariance\_vote\_mcq} & 90.0 & 37.3 \\
8  & composed  & \texttt{structural\_coordinate\_switch\_math} & 89.0 & 15.9 \\
9  & composed  & \texttt{terminal\_leakage\_sanitized\_math} & 90.0 & 17.4 \\
10 & new       & \texttt{first\_commitment\_race\_math} & 89.0 & 18.0 \\
\bottomrule
\end{tabular}
\caption{\textbf{Iteration-by-iteration progression on AQuA-RAT.} For each iteration, the candidate with the highest hidden-set accuracy and its weighted cost per sample; iteration 1 is the reference BL (CoT on \texttt{complex/thinking}). The final strategy (bold, from iteration 5) beats BL on both validation and hidden ($+5$ points at $0.77\times$ cost on hidden); the iteration-4 candidate, the most accurate on hidden, is not picked because on validation it matches BL's accuracy at $14\%$ higher cost.}
\label{tab:iteration_log}
\end{table}

\section{Per-Benchmark Result Tables}

\subsection{Per-Benchmark Main Results}
\label{app:main_table}

\tabref{tab:main_results} reports per-benchmark numbers behind \figref{fig:avg_bars} for the 30 main benchmarks. Cost is weighted tokens per sample (parameter-count weights), in thousands. Colors and arrows follow the rest of the paper.

\begin{table}[!htbp]
\centering
\footnotesize
\setlength{\tabcolsep}{6pt}
\begin{tabular}{lccccccc}
\toprule
Benchmark & Ours Acc & BL Acc & $\Delta$Acc & Ours Cost ($k$) & BL Cost ($k$) & $\Delta$Cost & Result \\
\midrule
SQuAD v2        & 79.1 & 73.2 & \textcolor{wincol}{$\uparrow$8.0\%} & 5.6 & 25.0 & \textcolor{wincol}{$\downarrow$77.4\%} & \textcolor{wincol}{Win} \\
TruthfulQA      & 69.0 & 64.5 & \textcolor{wincol}{$\uparrow$7.0\%} & 5.8 & 7.5 & \textcolor{wincol}{$\downarrow$21.9\%} & \textcolor{wincol}{Win} \\
2WikiMQA        & 76.5 & 72.5 & \textcolor{wincol}{$\uparrow$5.5\%} & 33.8 & 98.6 & \textcolor{wincol}{$\downarrow$65.7\%} & \textcolor{wincol}{Win} \\
Amazon-Rec      & 28.0 & 24.0 & \textcolor{wincol}{$\uparrow$16.7\%} & 45.6 & 70.3 & \textcolor{wincol}{$\downarrow$35.1\%} & \textcolor{wincol}{Win} \\
SAT-Math        & 98.2 & 94.5 & \textcolor{wincol}{$\uparrow$3.8\%} & 4.9 & 12.1 & \textcolor{wincol}{$\downarrow$59.4\%} & \textcolor{wincol}{Win} \\
LSAT-AR         & 81.0 & 77.6 & \textcolor{wincol}{$\uparrow$4.4\%} & 54.9 & 82.0 & \textcolor{wincol}{$\downarrow$33.0\%} & \textcolor{wincol}{Win} \\
LSAT-RC         & 81.5 & 78.5 & \textcolor{wincol}{$\uparrow$3.8\%} & 12.4 & 24.8 & \textcolor{wincol}{$\downarrow$49.9\%} & \textcolor{wincol}{Win} \\
NQ-Open         & 19.0 & 17.0 & \textcolor{tiecol}{$\uparrow$11.8\%} & 19.1 & 34.4 & \textcolor{wincol}{$\downarrow$44.5\%} & \textcolor{wincol}{Win} \\
LSAT-LR         & 82.0 & 80.5 & \textcolor{tiecol}{$\uparrow$1.9\%} & 65.2 & 98.1 & \textcolor{wincol}{$\downarrow$33.5\%} & \textcolor{wincol}{Win} \\
Gaokao-Math     & 94.5 & 93.5 & \textcolor{tiecol}{$\uparrow$1.1\%} & 89.7 & 111.3 & \textcolor{wincol}{$\downarrow$19.4\%} & \textcolor{wincol}{Win} \\
MuSiQue         & 42.0 & 41.0 & \textcolor{tiecol}{$\uparrow$2.4\%} & 67.6 & 73.0 & \textcolor{wincol}{$\downarrow$7.4\%} & \textcolor{wincol}{Win} \\
MMLU-STEM-Hard  & 88.0 & 87.0 & \textcolor{tiecol}{$\uparrow$1.1\%} & 11.9 & 35.2 & \textcolor{wincol}{$\downarrow$66.2\%} & \textcolor{wincol}{Win} \\
MMLU            & 79.5 & 78.5 & \textcolor{tiecol}{$\uparrow$1.3\%} & 10.2 & 96.3 & \textcolor{wincol}{$\downarrow$89.4\%} & \textcolor{wincol}{Win} \\
AQuA-RAT        & 87.0 & 86.5 & \textcolor{tiecol}{$\uparrow$0.6\%} & 17.5 & 19.4 & \textcolor{wincol}{$\downarrow$9.6\%} & \textcolor{wincol}{Win} \\
StrategyQA      & 70.0 & 69.5 & \textcolor{tiecol}{$\uparrow$0.7\%} & 3.0 & 3.7 & \textcolor{wincol}{$\downarrow$19.6\%} & \textcolor{wincol}{Win} \\
WinoGrande      & 77.5 & 77.5 & \textcolor{tiecol}{0.0\%} & 4.9 & 5.6 & \textcolor{wincol}{$\downarrow$11.8\%} & \textcolor{wincol}{Win} \\
ANLI-R3         & 69.5 & 69.5 & \textcolor{tiecol}{0.0\%} & 4.6 & 9.6 & \textcolor{wincol}{$\downarrow$51.7\%} & \textcolor{wincol}{Win} \\
SVAMP           & 93.0 & 93.5 & \textcolor{tiecol}{$\downarrow$0.5\%} & 1.8 & 2.0 & \textcolor{wincol}{$\downarrow$9.3\%} & \textcolor{wincol}{Win} \\
SciQ            & 97.5 & 98.0 & \textcolor{tiecol}{$\downarrow$0.5\%} & 1.2 & 3.7 & \textcolor{wincol}{$\downarrow$69.1\%} & \textcolor{wincol}{Win} \\
HellaSwag       & 70.5 & 71.0 & \textcolor{tiecol}{$\downarrow$0.7\%} & 9.0 & 22.9 & \textcolor{wincol}{$\downarrow$60.9\%} & \textcolor{wincol}{Win} \\
BBH             & 85.0 & 86.0 & \textcolor{tiecol}{$\downarrow$1.2\%} & 23.3 & 42.5 & \textcolor{wincol}{$\downarrow$45.2\%} & \textcolor{wincol}{Win} \\
CommonsenseQA   & 76.5 & 77.5 & \textcolor{tiecol}{$\downarrow$1.3\%} & 5.8 & 14.8 & \textcolor{wincol}{$\downarrow$60.8\%} & \textcolor{wincol}{Win} \\
GSM8K           & 92.0 & 94.0 & \textcolor{tiecol}{$\downarrow$2.1\%} & 8.3 & 62.8 & \textcolor{wincol}{$\downarrow$86.8\%} & \textcolor{wincol}{Win} \\
ASDiv           & 93.0 & 95.0 & \textcolor{tiecol}{$\downarrow$2.1\%} & 6.9 & 36.6 & \textcolor{wincol}{$\downarrow$81.1\%} & \textcolor{wincol}{Win} \\
MBPP+           & 80.5 & 79.0 & \textcolor{tiecol}{$\uparrow$1.9\%} & 108.5 & 109.0 & \textcolor{tiecol}{$\downarrow$0.4\%} & \textcolor{tiecol}{Tie} \\
MinervaM        & 38.5 & 39.0 & \textcolor{tiecol}{$\downarrow$1.3\%} & 28.2 & 28.7 & \textcolor{tiecol}{$\downarrow$1.5\%} & \textcolor{tiecol}{Tie} \\
MedQA           & 68.5 & 72.5 & \textcolor{losecol}{$\downarrow$5.5\%} & 12.6 & 28.5 & \textcolor{wincol}{$\downarrow$55.8\%} & \textcolor{losecol}{Mixed} \\
AIME 2024       & 60.0 & 65.0 & \textcolor{losecol}{$\downarrow$7.7\%} & 287.6 & 372.5 & \textcolor{wincol}{$\downarrow$22.8\%} & \textcolor{losecol}{Mixed} \\
HotpotQA        & 64.5 & 64.0 & \textcolor{tiecol}{$\uparrow$0.8\%} & 34.7 & 31.9 & \textcolor{losecol}{$\uparrow$9.0\%} & \textcolor{losecol}{Lose} \\
COPA            & 100.0 & 100.0 & \textcolor{tiecol}{0.0\%} & 2.4 & 2.3 & \textcolor{losecol}{$\uparrow$6.1\%} & \textcolor{losecol}{Lose} \\
\bottomrule
\end{tabular}
\caption{\textbf{Per-benchmark main results across 30 benchmarks.} BL is the baseline cell (of the $18\times6$ matrix) with the highest accuracy on the benchmark's training split, scored on the test set. $\Delta$ values are relative changes; colors follow the 2-point accuracy and 2\% cost tie bands. Result: Win = better on accuracy or cost and worse on neither; Mixed = better on one and worse on the other; Lose = worse on at least one and better on neither. Cost is weighted tokens per sample (parameter-count weights), in thousands.}
\label{tab:main_results}
\end{table}

\begin{figure}[!htbp]
\centering
\includegraphics[width=.9\linewidth, trim=0cm 1.5cm 0cm 0cm, clip]{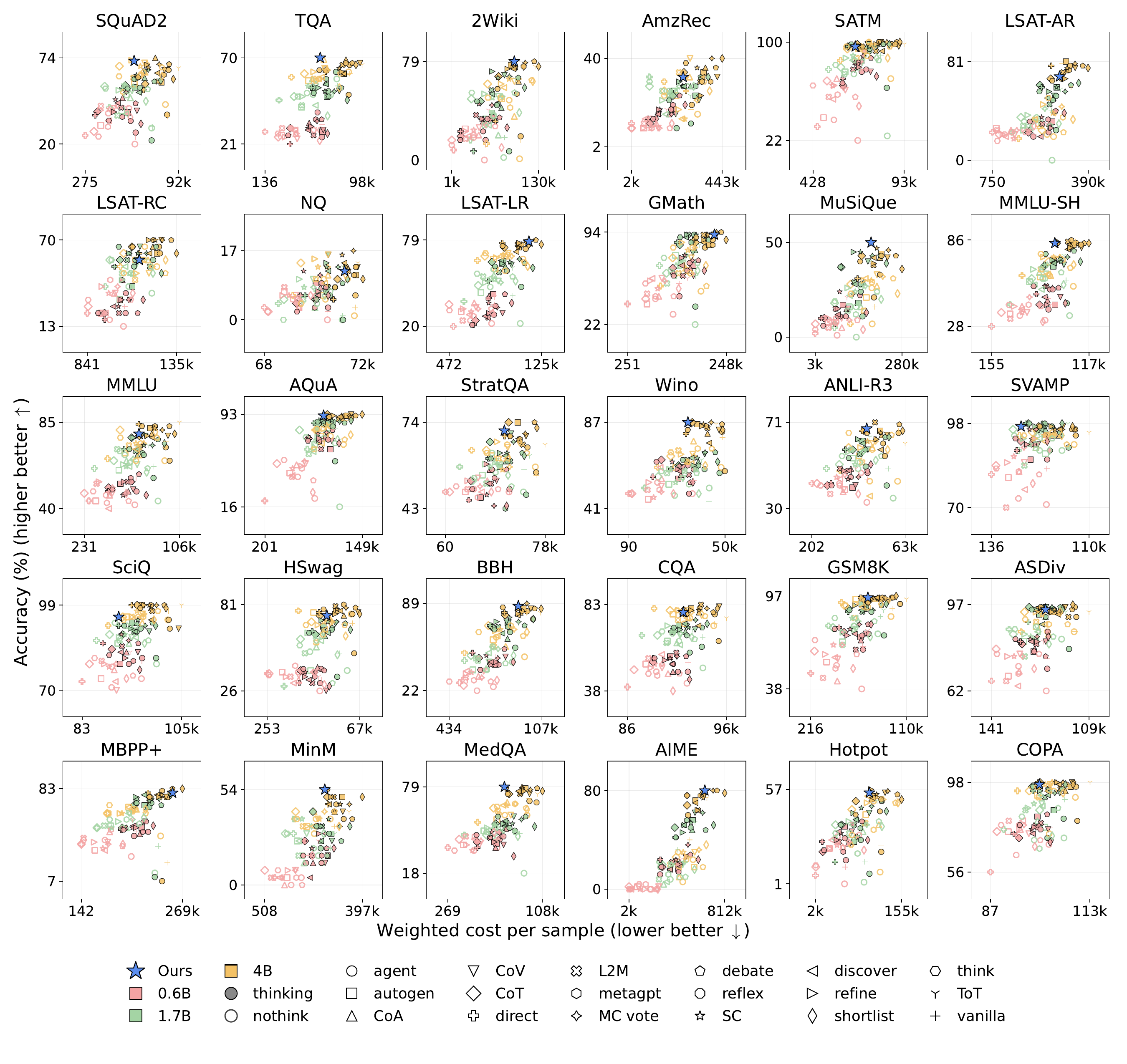}
\caption{\textbf{The same view on the training split.} As \figref{fig:cost_acc_main}, but every baseline cell and \method{}'s final strategy are scored on the benchmark's training split (the rows the strategist sees) instead of the test set. This comparison is biased toward BL: BL is, by definition, the most accurate of the 108 cells on exactly these rows, so it sits at the top of each panel, while \method{}'s strategy was chosen on the separate validation and hidden splits. Only the test set (\figref{fig:avg_bars}, \figref{fig:cost_acc_main}) scores both on questions neither was chosen on, which is why the paper reports it.}
\label{fig:cost_acc_train}
\end{figure}

\subsection{Per-Benchmark One-Shot Library Retrieval}
\label{app:oneshot_full}

\tabref{tab:oneshot_full} reports the absolute Ours/Baseline accuracy and cost values underlying \tabref{tab:oneshot}.

\begin{table}[!htbp]
\centering
\small
\setlength{\tabcolsep}{4pt}
\begin{tabular}{lcccccc}
\toprule
Benchmark & Ours Acc & Ours Cost ($k$) & BL Acc & BL Cost ($k$) & $\Delta$Acc & $\Delta$Cost \\
\midrule
ReCLor & 83.5 & 47.4 & 85.0 & 72.0 & \textcolor{tiecol}{$\downarrow$1.8\%} & \textcolor{wincol}{$\downarrow$34.2\%} \\
SAT-EN & 86.5 & 16.7 & 86.5 & 37.6 & \textcolor{tiecol}{0.0\%} & \textcolor{wincol}{$\downarrow$55.6\%} \\
TriviaQA & 50.5 & 25.8 & 46.0 & 64.6 & \textcolor{wincol}{$\uparrow$9.8\%} & \textcolor{wincol}{$\downarrow$60.0\%} \\
MATH500 & 94.5 & 47.2 & 92.5 & 61.0 & \textcolor{tiecol}{$\uparrow$2.2\%} & \textcolor{wincol}{$\downarrow$22.6\%} \\
BoolQ & 85.5 & 32.7 & 73.5 & 1.0 & \textcolor{wincol}{$\uparrow$16.3\%} & \textcolor{losecol}{$\uparrow$3103\%} \\
SIQA & 77.5 & 6.7 & 73.5 & 1.5 & \textcolor{wincol}{$\uparrow$5.4\%} & \textcolor{losecol}{$\uparrow$337\%} \\
OpenBookQA & 94.5 & 6.9 & 93.0 & 13.6 & \textcolor{tiecol}{$\uparrow$1.6\%} & \textcolor{wincol}{$\downarrow$49.2\%} \\
PIQA & 89.0 & 5.8 & 87.5 & 16.3 & \textcolor{tiecol}{$\uparrow$1.7\%} & \textcolor{wincol}{$\downarrow$64.5\%} \\
QUALITY & 79.5 & 52.0 & 78.5 & 48.4 & \textcolor{tiecol}{$\uparrow$1.3\%} & \textcolor{losecol}{$\uparrow$7.4\%} \\
DROP & 72.5 & 9.5 & 61.5 & 4.9 & \textcolor{wincol}{$\uparrow$17.9\%} & \textcolor{losecol}{$\uparrow$92.4\%} \\
\bottomrule
\end{tabular}
\caption{\textbf{Full per-benchmark one-shot library retrieval results.} Accuracy is on the test set; cost is weighted tokens per sample (parameter-count weights). $\Delta$Acc and $\Delta$Cost are relative changes vs the baseline (formula and color/arrow conventions match \tabref{tab:main_results}).}
\label{tab:oneshot_full}
\end{table}

\subsection{Per-Benchmark Cross-Model Transfer}
\label{app:transfer_full}

\tabref{tab:transfer_full} reports the absolute Ours/Baseline accuracy and total USD cost values underlying \tabref{tab:transfer}.

\begin{table}[!htbp]
\centering
\footnotesize
\setlength{\tabcolsep}{2.5pt}
\begin{tabular}{lcccccc@{\hspace{8pt}}cccccc}
\toprule
 & \multicolumn{6}{c}{GPT-5} & \multicolumn{6}{c}{Gemini-2.5} \\
\cmidrule(lr){2-7} \cmidrule(lr){8-13}
Benchmark & BL Acc & Ours Acc & $\Delta$Acc & BL \$ & Ours \$ & $\Delta$Cost & BL Acc & Ours Acc & $\Delta$Acc & BL \$ & Ours \$ & $\Delta$Cost \\
\midrule
AQuA-RAT & 31.0 & 89.5 & \textcolor{wincol}{$\uparrow$189\%} & 12.39 & 1.67 & \textcolor{wincol}{$\downarrow$86.5\%} & 71.0 & 90.5 & \textcolor{wincol}{$\uparrow$27.5\%} & 2.22 & 1.07 & \textcolor{wincol}{$\downarrow$51.6\%} \\
2WikiMQA & 29.5 & 78.5 & \textcolor{wincol}{$\uparrow$166\%} & 7.45 & 3.67 & \textcolor{wincol}{$\downarrow$50.8\%} & 21.5 & 77.5 & \textcolor{wincol}{$\uparrow$260\%} & 2.02 & 0.57 & \textcolor{wincol}{$\downarrow$72.0\%} \\
HotpotQA & 39.0 & 76.5 & \textcolor{wincol}{$\uparrow$96.2\%} & 7.71 & 7.13 & \textcolor{wincol}{$\downarrow$7.5\%} & 33.0 & 76.5 & \textcolor{wincol}{$\uparrow$132\%} & 2.52 & 1.20 & \textcolor{wincol}{$\downarrow$52.2\%} \\
MBPP+ & 56.5 & 82.5 & \textcolor{wincol}{$\uparrow$46.0\%} & 2.16 & 1.89 & \textcolor{wincol}{$\downarrow$12.3\%} & 69.5 & 80.5 & \textcolor{wincol}{$\uparrow$15.8\%} & 1.79 & 1.37 & \textcolor{wincol}{$\downarrow$23.7\%} \\
SQuAD2 & 57.5 & 74.0 & \textcolor{wincol}{$\uparrow$28.7\%} & 0.29 & 0.11 & \textcolor{wincol}{$\downarrow$61.5\%} & 74.0 & 70.5 & \textcolor{losecol}{$\downarrow$4.7\%} & 0.37 & 0.16 & \textcolor{wincol}{$\downarrow$57.9\%} \\
NQ-Open & 10.0 & 13.5 & \textcolor{wincol}{$\uparrow$35.0\%} & 9.28 & 7.17 & \textcolor{wincol}{$\downarrow$22.8\%} & 11.0 & 24.0 & \textcolor{wincol}{$\uparrow$118\%} & 1.91 & 1.50 & \textcolor{wincol}{$\downarrow$21.8\%} \\
TruthfulQA & 84.0 & 95.5 & \textcolor{wincol}{$\uparrow$13.7\%} & 3.86 & 1.08 & \textcolor{wincol}{$\downarrow$71.9\%} & 85.0 & 93.0 & \textcolor{wincol}{$\uparrow$9.4\%} & 1.31 & 0.50 & \textcolor{wincol}{$\downarrow$61.9\%} \\
ASDiv & 93.0 & 95.0 & \textcolor{tiecol}{$\uparrow$2.2\%} & 0.67 & 1.76 & \textcolor{losecol}{$\uparrow$163\%} & 93.5 & 95.5 & \textcolor{tiecol}{$\uparrow$2.1\%} & 0.88 & 0.59 & \textcolor{wincol}{$\downarrow$33.1\%} \\
HellaSwag & 93.5 & 94.0 & \textcolor{tiecol}{$\uparrow$0.5\%} & 0.07 & 0.09 & \textcolor{losecol}{$\uparrow$16.9\%} & 91.5 & 92.0 & \textcolor{tiecol}{$\uparrow$0.5\%} & 0.50 & 0.44 & \textcolor{wincol}{$\downarrow$12.3\%} \\
MinervaM & 24.0 & 24.5 & \textcolor{tiecol}{$\uparrow$2.1\%} & 9.97 & 10.45 & \textcolor{losecol}{$\uparrow$4.8\%} & 26.5 & 24.0 & \textcolor{losecol}{$\downarrow$9.4\%} & 4.96 & 3.68 & \textcolor{wincol}{$\downarrow$25.9\%} \\
\midrule
\textbf{Total} & & & & \textbf{53.86} & \textbf{35.02} & \textbf{\textcolor{wincol}{$\downarrow$35.0\%}} & & & & \textbf{18.50} & \textbf{11.07} & \textbf{\textcolor{wincol}{$\downarrow$40.1\%}} \\
\bottomrule
\end{tabular}
\caption{\textbf{Full per-benchmark cross-model transfer results.} For each benchmark, the \method{}-discovered strategy and the strongest leaderboard baseline workflow are both re-executed on the new backend; $\Delta$Acc and $\Delta$Cost compare them on the same backend. 200 deterministic samples per cell. BL \$ and Ours \$ are total USD across all 200 samples, computed at the official list prices published on the OpenAI (GPT-5) and Google (Gemini-2.5) websites; the same per-token rates are applied to baseline and \method{} strategies.}
\label{tab:transfer_full}
\end{table}

\section{Extended Related Work}
\label{app:related_extended}

This appendix expands the three Related Work paragraphs in the main text. Citations are identical; only the discussion is longer.

\paragraph{LLM agents.}
Large language models have rapidly evolved from single-call question answerers into multi-step agent systems. Chain-of-thought prompting~\citep{wei2022cot} and self-consistency decoding~\citep{wang2023selfconsistency} showed that intermediate reasoning steps improve accuracy on complex tasks. Tree of Thoughts~\citep{yao2023tot} extended this to deliberate search over reasoning paths. ReAct~\citep{yao2023react} interleaved reasoning with tool use, and Toolformer~\citep{schick2023toolformer} taught models to call external APIs autonomously. Reflexion~\citep{shinn2023reflexion} introduced verbal self-reflection as a feedback mechanism. Multi-agent frameworks such as AutoGen~\citep{wu2023autogen} enabled conversational collaboration among multiple LLM agents. Formal representations for these systems range from state machines~\citep{wu2024stateflow} to declarative compilation~\citep{khattab2024dspy} to optimizable computational graphs~\citep{zhuge2024gptswarm}. A recent survey~\citep{survey2025acg} organizes this landscape as agentic computation graphs, distinguishing static, pre-execution, and in-execution optimization. As these systems grow more complex, the strategy layer that determines how models are called, in what order, and with what verification has become the central design bottleneck. \method{} addresses this bottleneck by automating the design of the strategy layer itself.

\paragraph{Automated design of agentic systems.}
Structure search methods write strategies as raw code~\citep{hu2024adas} or compose them from fixed components~\citep{zhang2024aflow} or predefined modules~\citep{shang2024agentsquare, zhang2025evoflow}; the first yields isolated programs, the second bounds what strategies can be discovered. Optimization methods refine one strategy per task, by reinforcement learning~\citep{autoflow2025, zhang2026flowsteer, flowreasoner2025} or by LLM-driven evolution and refinement~\citep{gao2023strategyllm, sew2025}, and start over on the next task. Routing methods train controllers to select among existing strategies~\citep{arc2025, yue2025masrouter} without constructing them~\citep{bingreeky2025maas}. \method{} differs by allowing freeform creation through the Strategy Blueprint, requiring no training, constructing strategies itself, and routing across model tiers within each strategy.

\paragraph{Cross-task transfer and experience accumulation.}
Transferring knowledge across tasks has been studied through meta-learning~\citep{finn2017maml, hospedales2022metalearning}, neural architecture search~\citep{elsken2019nas}, and prompt optimization~\citep{yang2024opro}. In the LLM agent setting, ExpeL~\citep{zhao2024expel} extracts reusable insights from agent experiences, Voyager~\citep{wang2023voyager} grows a skill library as it explores new environments, and recent work has formalized this direction as lifelong learning for LLM agents~\citep{lifelong2025roadmap}. Despite this, automated system design methods~\citep{hu2024adas, zhang2024aflow} still optimize per task independently. \method{} maintains a shared library of blueprints and their measured properties that grows with every task, providing both relevant components (exploitation) and diverse inspiration from other tasks (exploration). To our knowledge, it is the first system for automated strategy design that accumulates knowledge across tasks.

\end{document}